\documentclass[11pt]{article}

\usepackage[T1]{fontenc}
\usepackage[utf8]{inputenc}
\usepackage[english]{babel}
\usepackage{lmodern}
\usepackage{amsmath,amsfonts,amssymb,amsthm}
\usepackage{graphicx}
\usepackage{booktabs}
\usepackage{multirow}
\usepackage{enumitem}
\usepackage{pifont}
\usepackage{authblk}
\usepackage[explicit]{titlesec}
\usepackage{xifthen}
\usepackage[noend]{algpseudocode}
\usepackage{algorithm}
\usepackage{microtype}
\usepackage{xurl}
\usepackage[numbers,sort&compress]{natbib}
\usepackage[colorlinks=true,allcolors=blue,bookmarks=true]{hyperref}
\usepackage{etoolbox}

\usepackage[letterpaper,margin=0.85in,top=0.6in,bottom=0.6in]{geometry}

\newcommand{\cmark}{\ding{51}}
\newcommand{\xmark}{\ding{55}}
\newcommand{\rank}{\operatorname{rank}}
\newcommand{\col}{\operatorname{col}}
\newcommand{\R}{\mathbb{R}}
\newcommand{\diag}{\operatorname{diag}}

\newtheorem{theorem}{Theorem}
\newtheorem{remark}{Remark}
\newtheorem{lemma}{Lemma}

\newtheorem{definition}{Definition}
\newtheorem{assumption}{Assumption}

\titleformat{\paragraph}[runin]
  {\sffamily\small\itshape\bfseries}
  {}
  {0em}
  {#1.}

\newboolean{shortarticle}
\newboolean{singlecolumn}
\setboolean{shortarticle}{false}
\setboolean{singlecolumn}{false}
\newcommand{\abscontentformatted}{}
\newcommand{\abscontent}{}
\newcommand{\Firstpage}{}
\newcommand{\Endparasplit}{}
\makeatletter
\def\ps@firststyle{\ps@plain}
\makeatother

\newcommand{\leadauthor}[1]{}
\newcommand{\dates}[1]{}
\newcommand{\doi}[1]{}
\newcommand{\keywords}[1]{\gdef\arxivkeywords{#1}}
\newcommand{\significancestatement}[1]{\gdef\arxivsig{#1}}
\newcommand{\authordeclaration}[1]{\gdef\arxivdeclare{#1}}
\newcommand{\equalauthors}[1]{\gdef\arxivequal{#1}}
\newcommand{\correspondingauthor}[1]{\gdef\arxivcorr{#1}}
\newcommand{\matmethods}[1]{\gdef\arxivmatmethods{#1}}
\newcommand{\acknow}[1]{\gdef\arxivacknow{#1}}
\def\arxivkeywords{}
\def\arxivsig{}
\def\arxivdeclare{}
\def\arxivequal{}
\def\arxivcorr{}
\def\arxivmatmethods{}
\def\arxivacknow{}

\def\arxivabstract{}
\long\def\abstract#1\end{%
  \long\gdef\arxivabstract{#1}%
  \end
}

\newcommand{\showmatmethods}{%
  \section*{Materials and Methods}
  \noindent\arxivmatmethods
}
\newcommand{\dataavail}[1]{%
  \par\vspace{0.9em}%
  \noindent\textbf{Data, Materials, and Software Availability.}\quad #1\par
}
\newcommand{\showacknow}{%
  \section*{Acknowledgments}
  \noindent\arxivacknow
}

\newcommand{\SItext}{%
  \clearpage
  \section*{Supporting Information}%
}

\makeatletter
\let\arxivbibliography\bibliography
\renewcommand{\bibliography}[1]{}
\makeatother

\date{}

\expandafter\let\csname figure*\endcsname\figure
\expandafter\let\csname endfigure*\endcsname\endfigure
\expandafter\let\csname table*\endcsname\table
\expandafter\let\csname endtable*\endcsname\endtable

\makeatletter
\patchcmd{\AB@maketitle}{\newpage}{\relax}{}{}
\patchcmd{\@maketitle}{\newpage}{\relax}{}{}
\let\arxivmaketitle\maketitle
\renewcommand{\maketitle}{%
  \g@addto@macro\@title{\thanks{This is a self-compiled version of the article published as: Qianli Liao, Liu Ziyin, Yulu Gan, Brian Cheung, Mark T. Harnett, and Tomaso Poggio. ``How biological synapses self-assemble gradient learning.'' \textit{Proceedings of the National Academy of Sciences} 123, no. 41 (2026): e2606245123. \url{https://doi.org/10.1073/pnas.2606245123.}}}%
  \arxivmaketitle
  \footnotetext[1]{Q.L., L.Z., and Y.G. contributed equally to this work.}%
  \footnotetext[2]{To whom correspondence should be addressed. E-mail: lql@mit.edu, tp@ai.mit.edu.}%
  \footnotetext[3]{Competing interests: Liu Ziyin was employed by NTT Research during the period when this work was carried out, and was previously employed by NTT (Nippon Telegraph and Telephone, Japan). The other authors declare no competing interests.}%
  \setcounter{footnote}{3}%
  \vspace{0.9em}
  \noindent\textbf{Abstract}\par
  \noindent\ignorespaces\arxivabstract\par
  \vspace{0.7em}
  \noindent\textbf{Significance}\par
  \noindent\arxivsig\par
  \vspace{0.55em}
  \noindent\textbf{Keywords:} \arxivkeywords\par
  \vspace{1.1em}
}
\makeatother

\hypersetup{%
  pdftitle={How Biological Synapses Self-Assemble Gradient Learning},
  pdfauthor={Qianli Liao, Liu Ziyin, Yulu Gan, Brian Cheung, Mark T. Harnett, Tomaso Poggio},
  pdfsubject={Authors' manuscript of Proc. Natl. Acad. Sci. U.S.A. 123 (41) e2606245123 (2026)}%
}

\begin{document}

\title{How Biological Synapses Self-Assemble Gradient Learning}

\author[a,b,c,1,2]{Qianli Liao}
\author[d,1]{Liu Ziyin}
\author[b,e,1]{Yulu Gan}
\author[b,e]{Brian Cheung}
\author[a,c,f]{Mark~T.~Harnett}
\author[a,b,c,e,f,g,2]{Tomaso~Poggio}

\affil[a]{K. Lisa Yang Integrative Computational Neuroscience Center, Yang Tan Collective, Massachusetts~Institute~of~Technology (MIT), Cambridge, MA 02139}
\affil[b]{Center for Brains, Minds, and Machines, MIT}
\affil[c]{McGovern Institute for Brain Research, MIT}
\affil[d]{Research Laboratory of Electronics, MIT}
\affil[e]{Computer Science and Artificial Intelligence Laboratory, MIT}
\affil[f]{Department of Brain and Cognitive Sciences, MIT}
\affil[g]{Italian Institute of Technology, Genova 16163, Italy}


\leadauthor{Liao}

\significancestatement{Learning in the brain is thought to emerge from local synaptic and circuit-level processes rather than from a globally designed blueprint. We formulate Self-Assembling Learning (SAL) as the problem of how such learning dynamics can arise from distributed, local interactions. We introduce a minimal circuit building block---the Self-Assembling Motif (SAM)---that can emerge from random connectivity under heterosynaptic plasticity rules and, when repeated across layers, self-organizes into networks that approximate stochastic gradient descent. This result shows that gradient-based optimization can arise from bottom-up self-organization, offering a principled link between biological plasticity and modern machine learning and suggesting testable predictions for cortical circuit organization.}

\authordeclaration{Competing interests: Liu Ziyin was employed by NTT Research during the period when this work was carried out, and was previously employed by NTT (Nippon Telegraph and Telephone, Japan). The other authors declare no competing interests.}
\equalauthors{\textsuperscript{1}Q.L., L.Z., and Y.G. contributed equally to this work.}
\correspondingauthor{\textsuperscript{2}To whom correspondence should be addressed. E-mail: lql@mit.edu, tp@ai.mit.edu
}

\keywords{Neural networks $|$ Biological learning $|$ Gradient descent $|$ Synaptic plasticity $|$ Self-assembly}

\begin{abstract} 
Existing models of learning in the brain explain how given circuits learn, but not how biology could assemble those circuits in the first place. To address this gap, we formulate Self-Assembling Learning (SAL)---the study of how learning systems can emerge from lower-level interactions---and introduce one example mechanism, the Self-Assembling Motif (SAM). SAM is self-assembling at two scales: the motif emerges from initially random connectivity under heterosynaptic plasticity rules, and networks of SAMs, composed hierarchically, self-organize into dynamics that provably approximate a generalized form of stochastic gradient descent---matching backpropagation-level performance. This suggests that biological learning need not be prescribed but can emerge from local rules---and to a far greater extent than previously thought.



\end{abstract}

\dates{This manuscript was compiled on \today}
\doi{\url{www.pnas.org/cgi/doi/10.1073/pnas.XXXXXXXXXX}}

\maketitle
\thispagestyle{firststyle}
\ifthenelse{\boolean{shortarticle}}{\ifthenelse{\boolean{singlecolumn}}{\abscontentformatted}{\abscontent}}{}

\Firstpage

\section{Introduction}

Deep neural networks have revolutionized computational neuroscience, where Artificial Neural Networks (ANNs) are increasingly used to model various aspects of information processing in the brain (\cite{cichy2016}; \cite{yamins2016}; \cite{guclu2015}; \cite{kriegeskorte2015}, \cite{lillicrap2020}; \cite{whittington2019}, \cite{sun2025}, \cite{raugel2025}). These models demonstrate remarkable success in capturing neural responses and cognitive functions. Yet a model's value as an account of the brain rests on more than reproducing such responses. A model of learning in the brain is \emph{biologically plausible} only if biology could actually produce it---a property we call biological \emph{realizability}. Realizability is not a verdict to be read off surface resemblance, nor a checklist of forbidden mechanisms to avoid; it is a demonstration that a mechanism could come to exist. And because biological structure has no external designer, any mechanism that realizes it must, at bottom, organize itself. Realizability is therefore fundamentally a question of self-organization---and, crucially, a matter of degree.

\begin{figure}[t!]
\centering
\includegraphics[width=\textwidth]{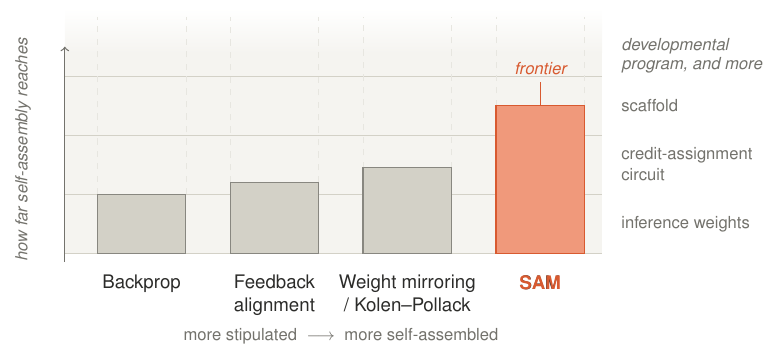}
\caption{\textbf{A model is biologically plausible only if biology could produce it on its own: biological plausibility/realizability as a graded axis of self-organization.} Learning algorithms differ in how much of their structure self-assembles from local interactions rather than being stipulated in advance. The vertical axis is an ordinal ladder of organizational levels: the inference/forward weights, the circuit that adjusts them, the scaffold that hosts it, and beyond. We label the second level the \emph{credit-assignment circuit} to be concrete, though more generally it is the \emph{learning circuit}---the machinery that turns experience into inference weight changes, of which credit assignment is the canonical example. Backprop and feedback-alignment variants self-organize essentially only their weights, leaving this circuit fixed; SAM assembles the circuit itself from uniform rules, resting only on a minimal two-stream scaffold whose form is largely required for stable perception. The faded top indicates that the axis is open-ended.}
\label{fig:axis}
\end{figure}

This reframing turns biological plausibility from a binary verdict into a graded axis (Figure~\ref{fig:axis}): learning models differ in how much of their structure is stipulated in advance versus shown to emerge from local interactions. The axis runs deep. One may ask only that the weights self-organize, while the circuit that adjusts them is designed; or, going further, that the learning circuit itself assemble; further still, the scaffold that hosts it; and in principle, the developmental program that lays the scaffold down. The further down a model reaches, the more of what others simply assume it must instead derive---to move down the axis is to replace assumption with assembly. By this measure, most existing learning algorithms sit very near the top: they self-organize their inference weights, but under machinery that is otherwise fixed in advance. This is true even when training yields rich, interpretable structure---reusable dynamical motifs~\cite{driscoll2024} or inhibitory microcircuits supporting working memory maintenance~\cite{kim2021}---because such structure emerges in the inference weights being trained, while the learning rule that adjusts them remains designed.

\Endparasplit

Nowhere is this stipulation more evident than in backpropagation, the standard
method for computing gradients in stochastic gradient descent (SGD). It
presupposes a precisely aligned architecture---forward and backward pathways tied
together by weight transport and neuron-level correspondence---that is difficult
to reconcile with biology \cite{liao2015, lillicrap2020, whittington2019}. 
Biologically plausible alternatives loosen parts of this machinery: feedback
alignment drops weight symmetry \cite{lillicrap2016,liao2015}, while weight
mirroring and the Kolen--Pollack rule let the backward weights adapt
\cite{akrout2019a}. Yet each still wires in a fixed one-to-one correspondence
between the two pathways rather than letting it form, and, more broadly, assumes
the prior existence of a specific learning-capable circuit that has not been
shown to be biologically assemblable
\cite{nokland2016, nokland2019, xiao2018, sacramento2018dendritic,
payeur2021burst, greedy2022single, max2024}. Such models explain how a fixed
circuit can learn, but not how the circuit itself arises---presupposing the very
structure whose origin is in question. We call this unresolved gap the
\emph{architectural circularity} of biologically plausible learning. A detailed
discussion of related approaches is provided in Section~\ref{sec:related}.

Stepping back from specific algorithmic proposals, we may instead ask a more
fundamental question: if biological learning systems are not globally designed,
what principles must govern their organization? Such a system must consist of
many interacting components, each following local rules, yet collectively giving
rise to coherent learning behavior at the system level. The problem is therefore
not to design a learning algorithm top-down, as in previous attempts, but to
understand how learning structure itself can emerge from distributed local
interactions.

This perspective leads us to formulate a scientific problem we call
\emph{Self-Assembling Learning} (SAL): how systems composed of interacting parts,
governed only by local interaction rules rather than centralized design, can
emerge from lower-level interactions into structures that implement effective
learning algorithms. Biological neural circuits provide a particularly rich and
concrete instantiation of this principle, though the SAL formulation is more
general and applies to any distributed system in which learning must emerge
rather than be imposed.

Addressing SAL requires bridging multiple levels of analysis. At the
computational level, one must specify what objective a self-organizing system
ultimately optimizes; at the algorithmic level, the dynamical process by which
parameters are updated; at the implementation level, the physical mechanisms
through which such dynamics are realized. These explanatory levels must in turn
connect to levels of biological organization---molecular processes, synaptic
interactions, circuit motifs, large-scale networks, and distributed systems.
These are, in effect, the same levels along which the self-organization axis
runs: to ask how far self-assembly reaches is to ask how many of these levels
emerge from local rules rather than being assumed.

The central challenge of SAL is therefore to understand how local mechanisms at
lower levels of organization can give rise to algorithmic learning dynamics at
higher levels, which in turn implement coherent computational objectives. In
neural systems, this means explaining how synaptic plasticity and cellular
dynamics can self-organize into circuits whose collective behavior resembles
principled optimization.

As a constructive step toward this goal, we introduce a circuit building block,
the \emph{Self-Assembling Motif} (SAM). SAM consists of four synaptic pathways
connecting two interacting information streams, all governed by uniform
heterosynaptic plasticity rules. SAM is self-assembling at two scales: first, the
motif itself emerges from initially random connectivity under local plasticity
rules; second, when repeated hierarchically across layers, networks of SAMs
self-organize into structured learning dynamics. In particular, the one-to-one
correspondence between pathways that prior methods wire in is, in SAM, not
assumed: its cross-stream connections are themselves randomly
initialized and plastic, so the alignment between the two streams forms on its own.

SAM is not assumption-free: it takes as given the existence of two
interacting streams with only modulatory---not driving---interactions between
them. This assumption is not arbitrary. Two segregated streams whose
cross-interactions cannot drive one another are close to what any system
performing stable perception requires: strong driving connections across streams
would form closed driving loops, leading to runaway amplification and the
contamination of inference by error signals. The two-stream scaffold is thus
better understood as a stability constraint that bounds the space of viable
architectures than as a designed blueprint, and it is exactly the kind of
low-information prior that development can plausibly supply. Within these minimal
constraints, everything in SAM that performs learning---the credit-assignment
circuit and all of its connections---self-assembles from initially random
connectivity under uniform local rules. SAM therefore sits further down the self-organization axis (Figure~\ref{fig:axis}) than any prior biologically plausible learning algorithm.
 
We theoretically prove that, up to a measurable approximation error, hierarchically composed SAM networks implement a generalized form of SGD whose effective learning rate is a matrix that itself emerges through plasticity and is asymptotically positive semidefinite. The emergence of such gradient dynamics from purely local rules shows
that optimization behavior can arise without top-down architectural
specification, making gradient-based learning compatible with biological
self-organization---and to a far greater extent than previously thought.

\section{Related Work}
\label{sec:related}
  
Our work is not the first to propose biologically plausible alternatives to backpropagation \cite{liao2015, lillicrap2016, nokland2016, xiao2018, nokland2019}. Recent efforts explore a variety of directions — top-down weight alignment \cite{max2024}, pseudoinverse feedback \cite{sejnowski2025}, and meta-learned plasticity rules over random feedback \cite{rosenbaum2023}, as well as objective-minimization accounts such as the free-energy principle \cite{friston2023}. The most closely related approach to ours is the Kolen-Pollack algorithm  \cite{kolen1994,akrout2019a}, a special case of our circuit in which the cross-connections are frozen and initialized to the identity. Very recently, \cite{abel2024} showed how learning can be integrated with other visual functions, such as visual guidance, by exploiting the interplay of ascending and descending cortical pathways. Below we examine several approaches in more detail, in two categories ordered by biological grounding. 

\subsection{ML-style biologically plausible learning}
A substantial body of work (see \cite{lillicrap2020, whittington2019} for comprehensive reviews) has sought to make gradient-based learning more biologically plausible by modifying some specific aspects of backpropagation while retaining most of its machine learning oriented designs. Feedback alignment and its variants replace exact weight symmetry with fixed and random feedback connections \cite{liao2015, lillicrap2016}.
Such approaches successfully eliminate the requirement of exact weight transport, but they typically rely on an artificial one-to-one correspondence between forward and backward neurons or pathways. The backward structure is assumed to mirror the forward structure and often derivatives of forward activations are re-used during the backward pass.

Direct feedback alignment and related methods \cite{nokland2016, nokland2019} further relax the architectural symmetry constraints in forward and backward pathways. However, they still keep the global constraints such as fixed and non-plastic feedback pathways, and the overall credit-assignment structure is specified a priori rather than emerging through biological assembly. Contrary to the feedback alignment philosophy, there is substantial evidence that biological synapses are highly plastic in feedback pathways during adulthood (see section \ref{sec:how_to_identify} for detailed discussion).

Finally, these models are typically formulated at the algorithmic level and do not provide detailed correspondence to specific biological cell types or microcircuit organization. 


\subsection{Models with stronger biological grounding}
More recent work has pursued tighter correspondence with cortical microcircuit structure by embedding credit assignment mechanisms directly within biologically grounded neuronal models. Examples include burst-based models such as BurstProp \cite{payeur2021burst} and BurstCCN \cite{greedy2022single}, as well as two-compartment neuron models such as Error-encoding Dendritic Networks (EDNs) \cite{sacramento2018dendritic} and \cite{max2024}.

Unlike feedback alignment methods, these approaches attempt to resolve the artificial one-to-one duplication of forward and backward neurons by allowing feedback signals to propagate through the original feedforward neurons themselves. We refer to this strategy as \textbf{original-pathway feedback}. Rather than introducing a separate backward population that mirrors the forward pathway, credit assignment signals are multiplexed onto the same neurons that transmit activations.

Two interesting mechanisms have been proposed within these models. In burst-based models, bursts are treated as distinct from ordinary spikes, enabling activation and error signals to coexist within the same neuronal pathway. This activation–error multiplexing is an elegant conceptual solution to the alignment problem. However, it assumes that bursts encode error-specific information, a hypothesis that remains to be firmly established biologically.

Alternatively, in models such as EDN, small perturbations or “nudges” are introduced to neuronal activations, allowing discrepancies to propagate across layers through multi-compartment dynamics. While this avoids explicit weight transport, the approach may be sensitive to other sources of activity fluctuation. Moreover, in its current formulation, EDN retains a one-to-one correspondence between interneurons in layer
$k$ and pyramidal neurons in layer $k+1$, requiring a somewhat artificial alignment at the circuit level.

Importantly, although these models achieve detailed biological correspondence at the cellular and dendritic levels, they generally assume the existence of a learning-capable microcircuit. The motif responsible for credit assignment is specified rather than shown to emerge through local assembly processes. As a result, they do not fully address the architectural circularity underlying the origin of learning-capable structure.

In contrast, our approach employs what may be termed \textbf{other-pathway feedback}, in which error signals propagate through distinct pathways that do not reuse the original feedforward neurons as intermediate carriers. This separation avoids activation–error multiplexing and does not require distinguishing bursts from spikes. More importantly, the feedback-supporting circuit motif itself is self-assembling under local plasticity rules, addressing both credit assignment and architectural emergence within a unified framework. Biology may employ either or both strategies, but here we focus on end-to-end self-assemblability of the learning architecture.

\section{Results}

\begin{figure*}[t!]
\centering
\includegraphics[width=\textwidth]
{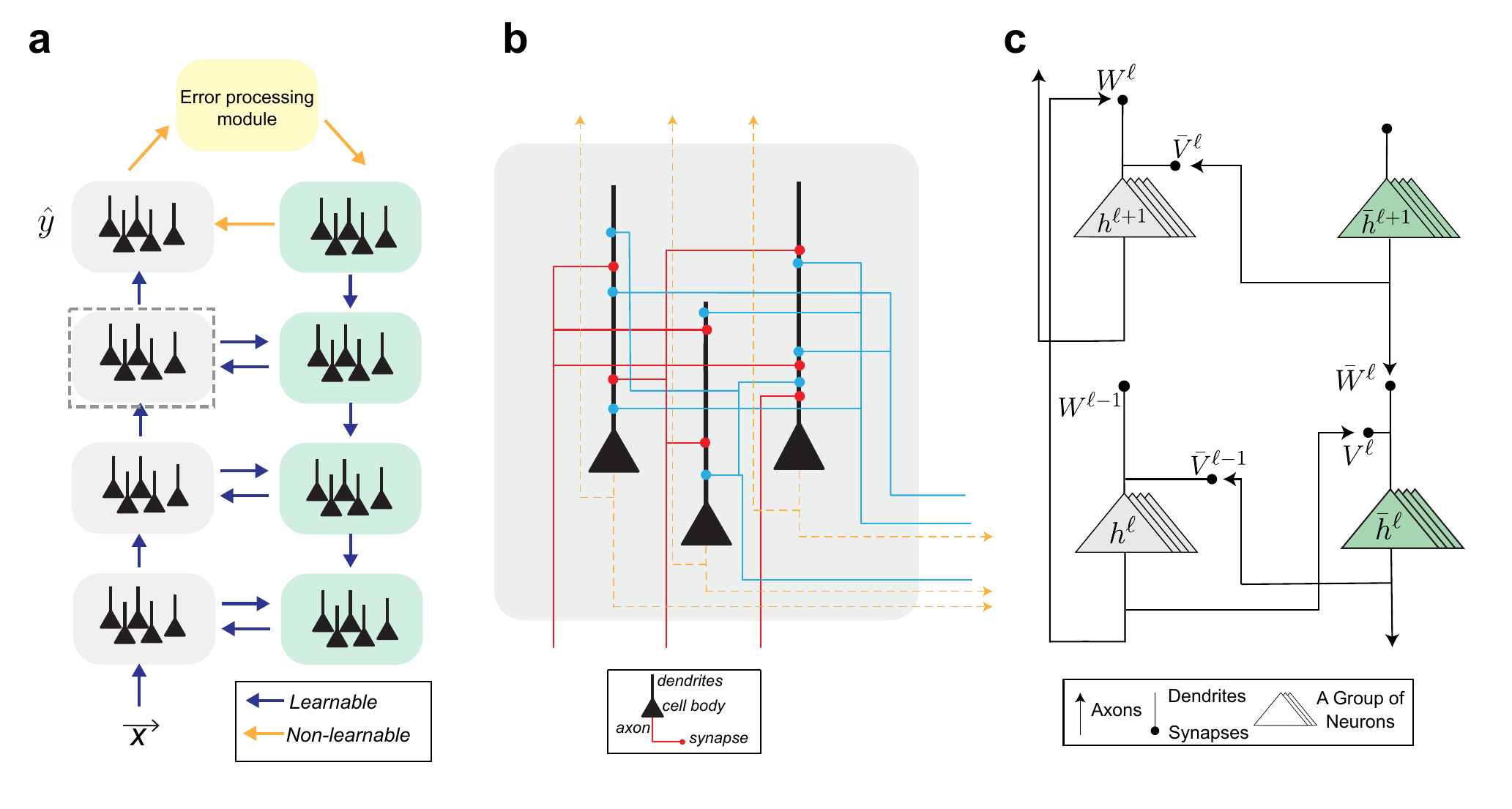}
    \caption{\small \textbf{A synaptic motif for interactions between ascending and descending information streams.} \textbf{(a)} Schematic of the upstream-downstream architecture. The upstream consists of a standard fully-connected neural network with multiple layers, roughly corresponding to a multi-region cortical processing pathway (e.g. V1-V2-V4-IT). The output of the upstream network goes to an error processing module (potentially corresponding to frontal cortices). This module computes a local error signal. This error signal is sent to the feedback (downstream) pathway, which processes information layer by layer downwards. The orange arrows represent non-learnable (identity) connections. Blue arrows represent learnable connections, each parameterized by a fully-connected weight matrix. \textbf{(b)} A detailed view of the smallest unit of the connection motif in panel a, indicated by grey dashed box. The neurons in panel b (as well as the abstract forms in A and C are illustrated as pyramidal neurons, the principal cell type of the cortex. Only apical dendrites are illustrated here for simplicity. \textbf{(c)} A mathematical description of the unit in panel b. Each arrow (which corresponds to a collection of axons, dendrites, and synapses) represents a set of full connections between two groups of neurons, parameterized by a weight matrix W or V . Every h in C is a vector and refers to the activations of a group of neurons. The connection matrix V allows upstream and downstream networks to have different number of hidden units in corresponding layers.}
    \label{fig: motif}
\end{figure*}

\subsection{A neural circuit architecture that self-assembles}

We propose a new ANN circuit architecture that is grounded in simple, biological synaptic motifs (Fig.~2). It is based on two interacting streams of information flow: an ascending stream representing forward pathways and a descending stream representing feedback pathways, mirroring cortical connectivity between areas. These streams are defined by a set of synaptic weights, which are subject to local heterosynaptic plasticity rules. Unlike the standard computer implementation of gradient descent – backpropagation – our model does not require the explicit computation of derivatives, and does not require weight symmetry between forward and feedback pathways.

There are a number of key features that differentiate our new ANNs from other approaches:
\begin{itemize}[noitemsep,topsep=0pt, parsep=0pt,partopsep=0pt, leftmargin=13pt]
\item {\it \textbf{Heterosynaptic Learning Rules}}: All synaptic updates are driven by the activity of the synapse itself and of other synapses on the same neuron. The latter do not need to be electrically close to the modifiable synapse, depending on the specific mechanism of heterosynaptic plasticity.
\item {\it \textbf{Biological Congruence}}: The architecture aligns with known anatomical and physiological properties of cortical circuits.
\item {\it \textbf{Reduced Assumptions}}: Our models do not require any weight symmetry and still achieve competitive performance on challenging datasets like CIFAR-10;
\item {\it \textbf{Structural Resilience}}: Our circuit motif works with a range of nonlinearities underlying the transduction from synaptic activities to neuron activity (and corresponding to the ReLU nonlinearity used in many neural networks) and for an arbitrary width of the forward and backward pathway. The network is therefore robust to small changes in the local configurations of pathways.
\end{itemize}

\paragraph{Notation} Let $\theta$ be the parameters of the model. We will use $\Delta f(\theta)$ to denote the difference of the quantity $f(\theta)$ after one step of the learning algorithm: $\Delta f(\theta) := f(\theta_{t+1}) - f(\theta_t)$. Any vector without a transpose superscript is treated as a column vector (e.g. $\nabla F(\theta)$ is a column vector for a scalar function $F$).

We assume that each pathway is organized in layers indexed with $\ell \in \{0,...,L\}$, with input at $0$ and readout at $L$ (depth $L$ in weight hops/modules; motifs $\ell=0,\ldots,L-1$). Let $h^\ell \in \mathbb{R}^{d_{\ell}}$ denote the neuron's cell body electric potential (i.e., before applying nonlinearity) of the $\ell$-th layer upstream pathway and $\bar{h}^{\ell} \in \mathbb{R}^{\bar{d}_{\ell}}$ the downstream pathway. For reasons that will become clear, the upstream is also referred to as the forward pathway, and the downstream is the feedback pathway. Note that it is, in general, the case that the two pathways have different numbers of neurons, and so $d_{\ell}$ is not necessarily equal to $\bar{d}_{\ell}$. The simplest such biological motif is shown in Figure~\ref{fig: motif}, where four synaptic connection matrices connect two consecutive layers of the two pathways. Here, $W^{\ell} \in \mathbb{R}^{d_{\ell+1} \times d_{\ell}}$ is the connection from the $\ell$-th layer upstream pathway to the $\ell+1$-th layer upstream pathway, $\bar{W}^\ell \in \mathbb{R}^{\bar{d}_{\ell} \times \bar{d}_{\ell+1}}$ the connection from $\ell+1$-th layer downstream to the $\ell$-th layer downstream. $V^{\ell} \in \mathbb{R}^{\bar{d}_{\ell} \times {d}_{\ell}}$ and $\bar{V}^\ell \in \mathbb{R}^{d_{\ell+1} \times \bar{d}_{\ell+1}}$ are inter-stream connections: $V$ goes from the upstream to the downstream, and $\bar{V}$ from the downstream to the upstream. The algorithm pseudocode is in SI Pseudo Code.

The computation of the two pathways takes place sequentially, where the upstream computation is performed first and eventually used for inference, and the downstream computation takes place after inference and is used for synaptic weight updates:\footnote{This sequential ordering is consistent with, for example, the observation that inactivating V1 also leads to the inactivation of V2, while deactivating V2 has no effect on the activations of V1 (\cite{anderson2009}).}
$h^{\ell+1} =  W^\ell D_u^\ell h^{\ell}$, $\bar{h}^{\ell} =\bar{W}^\ell  D_d^{\ell+1}  \bar{h}^{\ell+1}$, 
where $D$ matrices are diagonal matrices that function as the neuron nonlinearity and may be different for upstream ($D_u$, corresponding to $h$) and downstream ($D_d$, corresponding to $\bar{h}$) pathways and for different layers.  The input layer is $h^0 := x$, and $h^L := \hat{y}$ is the output of the network. The input to the downstream pathway is an error signal obtained from an objective function $F$: $\bar{h}^L := \epsilon(\hat{y}) = -\nabla_{\hat{y}} F(\hat{y})$, which is computed from the output of the upstream pathway. The downstream propagates this signal in the reverse direction. One example of $F$ is the MSE loss, where $F(\hat{y})=\|y-\hat{y}\|^2$, where $y$ is the correct label.



For the first layer, $D$ is identity (i.e., no nonlinearity on the input). For later layers, we choose $D_u$ to be a diagonal zero-one matrix:
$ D_u^\ell = {\rm diag}(\mathbf{1}_{(h^{\ell})_1},..., \mathbf{1}_{(h^{\ell})_d}) $
where $\mathbf{1}_x =1$ if $x>0$ and is zero if $x\leq 0$,  corresponding to a ReLU nonlinearity. For $D_d$, we focus on the simplest case where $D_d=I$ is the identity matrix.



Another common choice of $D_d$ is the Jacobian of a ReLU nonlinearity based on forward activations, which requires $V=\bar{V}=I$ (namely a one-to-one correspondence between upstream and downstream neurons). With this choice of $D_d$, $V$ and $\bar{V}$, our algorithm reduces to:
1. standard backpropagation, if $\bar{W}$ is hardwired to $W^T$ throughout training. 2. Feedback Alignment (\cite{lillicrap2016}), if $\bar{W}$ is random and fixed. 3. the KP-algorithm (\cite{kolen1994}), if $\bar{W}$ is learned using our learning rule, which converges to the exact gradient descent provably.  

\textbf{Heterosynaptic Learning:} our local learning rule takes the general form of $\Delta W \propto S_{post}*S_{pre}$, where the change $\Delta W$ of a synapse is proportional to the product of  postsynaptic signal $S_{post}$ and presynaptic signal $S_{pre}$. This is similar to Hebbian learning, but we have extra emphasis that $S_{post}$ refers to the signals sent by feedback synapses to the postsynaptic neuron instead of those arising from self-activation due to forward synapses. Feedback signals are informative in learning while self-activation signals only lead to non-informative arbitrary self-strengthening. 

In most situations, $S_{pre}$ represents the presynaptic potential on an axon, mathematically taking the form of $D_{pre}*h_{pre}$, where $D_{pre}$ is the nonlinearity and $h_{pre}$ is the electric potential of the presynaptic cell body. For $W$ and $\bar{W}$, the postsynaptic factor always includes a gain $D_{post}$; for $V$ and $\bar{V}$, whether the postsynaptic factor is gated ($S_{post}\propto D_{post}h_{post}$) or left ungated ($S_{post}\propto h_{post}$) is the main algorithmic variant, as noted below. 


This simple rule guides the learning of all four directions of synapses in our model:
\begin{flalign}\label{eq: sal start}
    &\Delta W^\ell  =\eta_W (D_u^{\ell+1}(\bar{V}^\ell D_d^{\ell+1}\bar{h}^{\ell+1})) ( D_u^\ell h^\ell)^T - \gamma W^\ell \nonumber \\ &\propto  (D_{Wpost}S_{Wpost}) ( D_{Wpre}S_{Wpre})^T ,\\[10pt]
    &\Delta \bar{W}^\ell  = \eta_{\bar{W}} (D_d^{\ell} (V^\ell D_u^{\ell} h^\ell)) ( D_d^{\ell+ 1} \bar{h}^{\ell+1})^T   - \gamma \bar{W}^\ell \nonumber \\ &\propto (D_{\bar{W}post} S_{\bar{W}post}) ( D_{\bar{W}pre} S_{\bar{W}pre})^T ,\\[10pt]
    &\Delta V^\ell  =  \eta_V (D_d^{\ell}\bar{h}^{\ell}) (D_u^{\ell}h^\ell)^T  - \gamma V^\ell  \nonumber\\ & \propto (D_{Vpost}S_{Vpost}) (D_{Vpre}S_{Vpre})^T,\\[10pt]
    &\Delta \bar{V}^\ell = \eta_{\bar{V}} (D_u^{\ell+1}h^{\ell+1})(D_d^{\ell+1}\bar{h}^{\ell+1})^T  - \gamma \bar{V}^\ell \nonumber\\ & \propto (D_{\bar{V}post}S_{\bar{V}post}) (D_{\bar{V}pre}S_{\bar{V}pre})^T, 
\end{flalign}
where $\eta$ is the time constant of learning (i.e., the learning rate), $\gamma$ is the strength of local regularization terms that encourage the synapses to be weak when unused. Here $S$ denotes a general signal that may or may not already be gated (as $S_{pre}$ usually is); gating a gated signal changes nothing, as $D^2=D$ for ReLU. This lets all four rules be written in the same symmetric gated form. The original benchmark experiments were run with the slightly simpler ungated interstream variant, which omits the postsynaptic nonlinearity on $V$ and $\bar{V}$ updates---equivalently $D_{V\mathrm{post}}=D_{\bar{V}\mathrm{post}}=I$---as in SI Pseudo Code, Algorithm~1. We did not observe qualitative differences between the two variants in practice. The alignment diagnostics and SI Theory use the gated form, SI Pseudo Code, Algorithm~2, because it is the more symmetric presentation; the same interstream-alignment mechanism applies to either form when the paired $V^\ell$ and $(\bar V^{\ell-1})^T$ updates use the same outer-product drive and matched learning rates. The four sets of rules are almost completely symmetric along all directions -- the only asymmetry between $W$/$\bar{W}$ and $V$/$\bar{V}$ stems from the fact that $W$ and $\bar{W}$ have the overwriting effect on the neurons that they connect to, while $V$ and $\bar{V}$ only serve to provide learning signals but do not overwrite the target neurons' activations. This design choice may be explored further in the future. We further discuss the biological plausibility of our learning rules in SI Biological Evidence for Synaptic Plasticity Rules.  



Both the learning rate and the regularization strength are labeled with a different subscript to emphasize that they could have different time constants. In this work, we always keep the regularization strengths uniform. We empirically search over 64 combinations of the learning rates and use the best one in the experiment.\footnote{See SI Experimental Setup.} We found it particularly beneficial to fix the ratio between the four learning rates, while tuning the overall factor for different tasks.

A key feature of this algorithm is that it allows both the inference network and the learning circuit, in addition to all their connections, to be simultaneously self-assembled. 

\begin{figure*}[t!]
\centering
    \includegraphics[width=\textwidth]{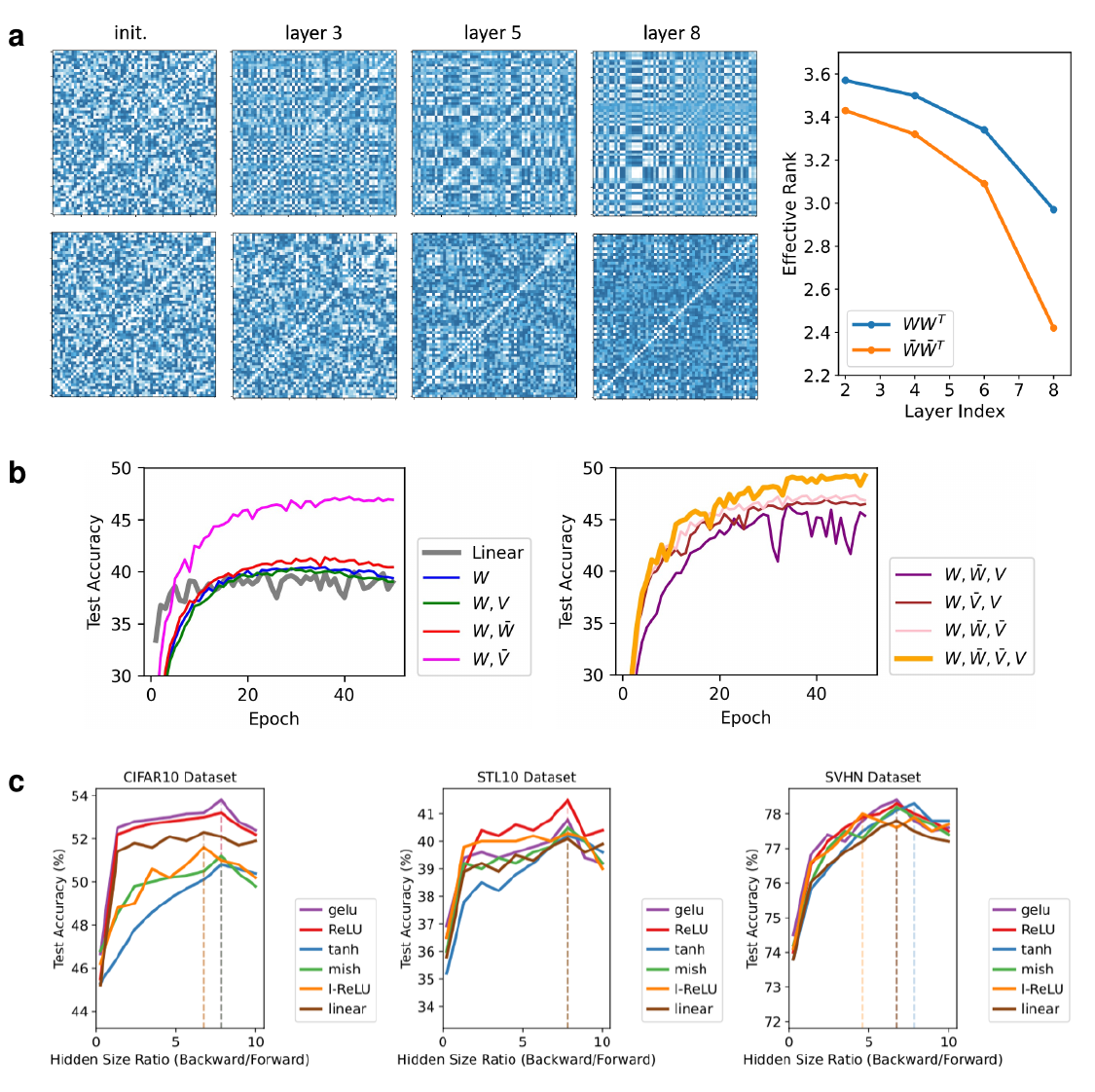}
    \caption{\textbf{Detailed analyses, variants and ablation studies of SAM.} \textbf{(a)} The matrices $\bar{V}\bar{V}^T$ (upper) and $WW^T$ (lower) before and after training. Also, recall that $\bar{V}\bar{V}^T = H^\ell$ determines the ideal aligned matrix learning rate $H_D^\ell$ for $W^\ell$ at $E_v^{\ell+1}=0$ (Theorem~\ref{theo:main}; more generally the prefactor is $H_D^\ell+\widetilde E_{v,D}^\ell$). The neurons within the same layer become correlated after training. The spectra of the weight matrices of the two pathways are found to be effectively low-rank (right). This visualization used the SVHN dataset with a network having 8 hidden layers. \textbf{(b)} Ablation study on roles of $\bar{V}$, $\bar{W}$ and $V$ on CIFAR-10. Here, we make a subset of the interconnections nontrainable. The legend labels indicate trainable components. ``Linear'' refers to a linear network trained with SGD. We see that (1) making $\bar{V}$ plastic is more important than making other connections plastic, and (2) making everything plastic significantly improves the performance further. \textbf{(c)} The impact of the width of the downstream pathway on performance. We use different activation functions in the forward pathway. The width of the upstream pathway was kept constant at 200, while varying the width of the downstream pathway. We find that having a wider downstream network improves the performance of the feedforward network up to an overparametrization ratio of 7.5. In biology, the ratios can also depend on biological constraints such as energy consumption or wiring volume and may be region-dependent.}
    \label{fig:detailed_analyses}
\end{figure*}

\subsection{Gradient Learning in Overparametrized Downstream Pathway}

The circuits we describe can be viewed as a generalized version of SGD, with a matrix-form learnable learning rate---the algorithm learns the learning algorithm itself.
The formal argument is organized around \textbf{pathway alignment} between the autograd route and the SAM feedback route at each layer.

\begin{figure*}[t!]
\centering
\includegraphics[width=\textwidth]{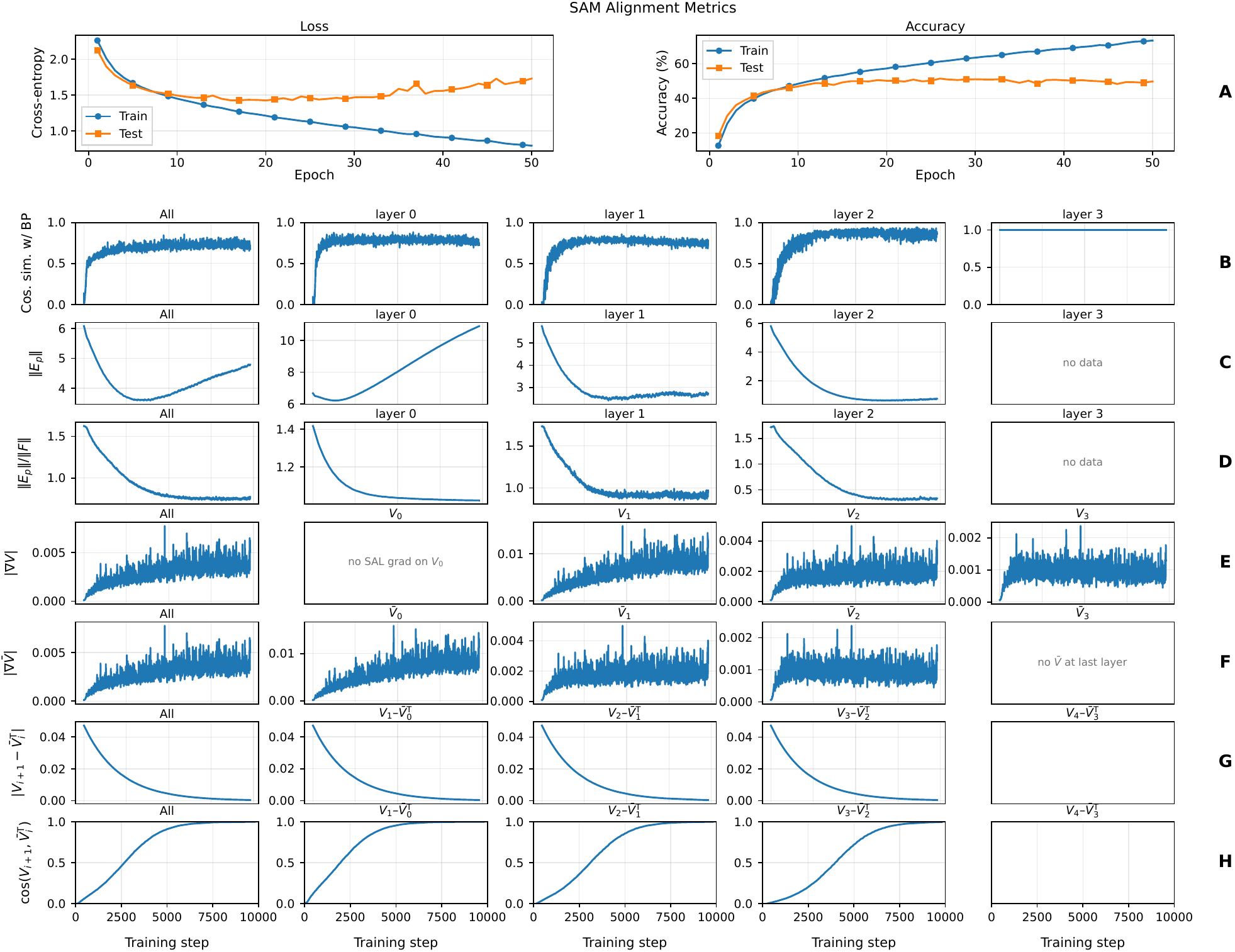}
\caption{\textbf{Alignment diagnostics on CIFAR-10 (3 hidden layers; stable run).}
SAL vs.\ backprop during a $50$-epoch run (hyperparameters in SI Alignment Diagnostics).
\textbf{A:} train and test cross-entropy and accuracy over epochs.
\textbf{B:} cosine similarities between SAL and backprop gradients on $W^\ell$ increase during learning and reach high level.
\textbf{C--D:} pathway error $\|E_p^\ell\|$ and $\|E_p^\ell\|/\|\mathcal{F}^\ell\|$ ($E_p^\ell=\mathcal{F}^\ell-\mathcal{B}^\ell$, normalized by $\|\mathcal{F}^\ell\|$); both decrease on most layers; $E_p$ tracks pathway misalignment and dominates the residual once $E_v$ is small.
\textbf{E--F:} interstream gradient magnitudes $|\nabla V^\ell|$ and $|\nabla\bar V^\ell|$ remain at a stable positive level.
\textbf{G--H:} interstream mismatch $|V^{\ell+1}-\bar V^{\ell\mathrm{T}}|$ falls and junction cosine similarity approaches one, consistent with unconditional $E_v^\ell\to 0$ even while interstream plasticity stays active.
Together, tight interstream alignment and decreasing pathway error match the conditional main theorem and help explain why SAL gradients track backprop (panel~B). These are early diagnostics; a systematic analysis of $E_p$ is left to future work.}
\label{fig:alignment-3layer}
\end{figure*}

\paragraph{Alignment errors}
Fix layer $\ell$ and sample $x$.
Let $g^{\ell+1}:=\partial F/\partial h^{\ell+1}$ denote the backprop signal.
Define the \textbf{forward pathway operator} $\mathcal{F}^\ell := V^\ell D_u^\ell (W^\ell)^T$ and the \textbf{feedback pathway operator} $\mathcal{B}^\ell := D_d^\ell\bar W^\ell(\bar V^\ell)^T$, both mapping $\mathbb{R}^{d_{\ell+1}}\to\mathbb{R}^{\bar d_\ell}$.
The \textbf{pathway alignment error} is
\begin{equation}\label{eq:Ep-main}
    E_p^\ell := \mathcal{F}^\ell - \mathcal{B}^\ell
    = V^\ell D_u^\ell (W^\ell)^T - D_d^\ell\bar W^\ell(\bar V^\ell)^T;
\end{equation}
ideal pathway alignment means $E_p^\ell=0$.
The \textbf{interstream alignment error} is
\begin{equation}\label{eq:Ev-main}
    E_v^\ell := V^\ell - (\bar V^{\ell-1})^T;
\end{equation}   
ideal interstream alignment means $E_v^\ell=0$ (equivalently $\bar V^{\ell-1}=(V^\ell)^T$).
When $\eta_V=\eta_{\bar V}$, the interstream updates imply $\Delta E_v^\ell=-\gamma E_v^\ell$, so $E_v^\ell\to 0$ exponentially from any initialization (SI Theory, Lemma~1).
\textbf{Interstream alignment} requires only that the misalignment operator $E_v^\ell$ tighten---driving $\bar V^{\ell-1}\to(V^\ell)^T$---not that $\Delta V^\ell$ or $\Delta\bar V^\ell$ vanish during training.
By contrast, $E_p^\ell$ is \textbf{pathway alignment} (\eqref{eq:Ep-main}): convergence of $E_p^\ell\to 0$ is \emph{not} proved from the learning rules alone, but in successful runs we typically observe $\|E_p^\ell\|$ decrease on most layers.
Because $E_p^\ell=\mathcal{F}^\ell-\mathcal{B}^\ell$ is the difference between the forward and feedback pathway operators, we also report $\|E_p^\ell\|/\|\mathcal{F}^\ell\|$: this asks how large that operator mismatch is relative to the forward pathway itself, giving a scale-free measure of whether $\mathcal{B}^\ell\approx\mathcal{F}^\ell$ that is comparable across layers (Figure~\ref{fig:alignment-3layer}, panels~C--D).
Theorem~\ref{theo:main} requires only that these errors stay small enough that the residual $R^\ell$ is negligible; when they instead grow---as at diverging layers in deeper networks---$E_p$ becomes the dominant practical source of alignment error.
We formalize the resulting weight update below; the full proof and explicit residual bounds in terms of $\|E_p^k\|$ and $\|E_v^k\|$ are given in SI Theory.

\begin{theorem}\label{theo:main}
    Consider a ReLU neural network in which each $V^\ell\in\mathbb{R}^{\bar d_\ell\times d_\ell}$ has full column rank (so $\bar d_\ell\ge d_\ell$; a wider downstream is allowed).
    Fix sample $x$ and layer $\ell$.
    Suppose $(\bar V^{L-1})^T=I$, $\eta_V=\eta_{\bar V}$,
    $\|E_p^k(x)\|\le\epsilon_p^k(x)$ for $k=\ell+1,\ldots,L-1$,
    and $\|E_v^k\|\le\epsilon_v^k$ for $k=\ell+1,\ldots,L-1$.
    Then
    \begin{equation}\label{eq:main-SAL}
        \Delta_{\rm SAL} W^\ell
        = \eta_W\,\bigl(H_D^\ell+\widetilde E_{v,D}^\ell\bigr)\,
        \Delta_{\rm sgd} W^\ell
        - \gamma W^\ell + R^\ell(x),
    \end{equation}
    where $H^\ell = \bar{V}^\ell (\bar{V}^\ell)^T$, $\widetilde E_v^\ell:=\bar V^\ell E_v^{\ell+1}$,
    \[
        H_D^\ell:=D_u^{\ell+1}H^\ell D_u^{\ell+1},
        \qquad
        \widetilde E_{v,D}^\ell:=D_u^{\ell+1}\widetilde E_v^\ell D_u^{\ell+1},
    \]
    $\Delta_{\rm sgd} W^\ell = -\nabla_{W^\ell} F$, and $R^\ell(x)$ is a remainder term (bounded in SI Theory).
    When $E_p^k=E_v^k=0$ on the chain, $R^\ell=0$, $\widetilde E_{v,D}^\ell=0$, and $\Delta_{\rm SAL} W^\ell = \eta_W H_D^\ell \Delta_{\rm sgd} W^\ell - \gamma W^\ell$. Since $E_v^\ell\to 0$ when $\eta_V=\eta_{\bar V}$ (SI Theory, Lemma~1), $\widetilde E_{v,D}^\ell$ decays during training and the prefactor approaches $H_D^\ell$, which is positive semidefinite.
\end{theorem}

\begin{remark}\normalfont
    Theorem~\ref{theo:main} separates two alignment notions.
    \textbf{Interstream alignment} ($E_v^\ell\to 0$) is proved unconditionally when $\eta_V=\eta_{\bar V}$: the streams become transpose-matched at each splice, $\bar V^{\ell-1}\to(V^\ell)^T$.
    \textbf{Pathway alignment} ($E_p^\ell$ small) is the conditional hypothesis under which the dual-stream motif implements generalized SGD with matrix learning rate $H_D^\ell+\widetilde E_{v,D}^\ell$.
    Empirically $\|E_p^\ell\|$ and $\|E_p^\ell\|/\|\mathcal{F}^\ell\|$ often decrease during stable training, but this is not proved from the rules alone; when and why $E_p^\ell$ improves or instead increases (as in diverging runs) is left to future work (Figure~\ref{fig:alignment-3layer}).
    Early in training, finite $\epsilon_v^k$ inflates the depth-chain residual before $E_v$ decays; as $\widetilde E_{v,D}^\ell$ and $R^\ell$ shrink, the update approaches SGD preconditioned by the learned matrix $H_D^\ell$, which is positive semidefinite and therefore does not oppose the SGD gradient.
\end{remark}
 
It is worth noticing that this algorithm is different from SGD. The effective learning rate on $W^\ell$ is the matrix $H_D^\ell+\widetilde E_{v,D}^\ell$, not a scalar: it is explicitly time-dependent and is itself learned by the algorithm. In the aligned limit this reduces to the positive semidefinite matrix $H_D^\ell$. This is consistent with observed metaplasticity in biological systems (\cite{abraham1996}; \cite{abraham2008}) -- though it extends the idea in a fundamental way since the learning algorithm itself is being learned during the training of the forward network. In Figure~3a, we show examples of the $H$ matrix after training.

Another interesting aspect of the algorithm (see Theorem~\ref{theo:main}) is that it allows (but does not require) the backward pathway to be overparametrized. That is, the number of downstream connections can be larger than the number of forward connections. This matches observations in biological circuits (\cite{briggs2020}), in which feedback connections can greatly outnumber feedforward ones---a flexibility that most prior biologically inspired learning algorithms do not accommodate because they require matched pathway sizes. Also, the backward pathway is flexible in the choice of $D_d$; the alignment framework and Theorem~\ref{theo:main} apply for the $D_d=I$ case used here and more generally.

\textbf{When does the theorem apply in practice?}
Interstream alignment is automatic when $\eta_V=\eta_{\bar V}$: $E_v^\ell$ decays exponentially (SI Theory, Lemma~1).
Pathway alignment is separate: $E_p^\ell=\mathcal{F}^\ell-\mathcal{B}^\ell$ directly measures whether the feedback pathway tracks the forward Jacobian.
In stable training, per-layer $\|E_p^\ell\|$ and $\|E_p^\ell\|/\|\mathcal{F}^\ell\|$ typically decrease; when a run diverges, a sharp rise in $\|E_p^\ell\|$ at an early layer often accompanies instability (SI Alignment Diagnostics, 4-hidden-layer example).
Theorem~\ref{theo:main} requires only that these quantities stay small enough that $R^\ell$ is negligible---and empirically, once interstream alignment has tightened, $E_p$ is the main contributor to the gradient-approximation residual. 
Two limiting cases are instructive: if $V^\ell$ and $\bar V^\ell$ are aligned permutation matrices and $D_d$ matches the forward nonlinearity, then $H^\ell=I$, $\widetilde E_v^\ell=0$, and SAM reduces to SGD; if the downstream pathway is wide, $\mathcal{B}^\ell$ has more capacity, which may help keep $\|E_p^\ell\|$ decreasing during training.
Simulations report per-layer $\|E_p^\ell(x)\|_F$ from \eqref{eq:Ep-main} with sample-dependent ReLU gates $D_u^\ell$ and $D_d^\ell=I$ (the benchmark setting; SI Alignment Diagnostics).

Figure~\ref{fig:alignment-3layer} tests this alignment framework on CIFAR-10 for a stable three-hidden-layer run; panel-by-panel readout is in the caption.
Full hyperparameters and a deliberately chosen divergent 4-hidden-layer illustration are in SI Alignment Diagnostics.

\begin{table*}[t!]
\centering
  \footnotesize
  \caption{\small \textbf{Performance of the proposed method compared with SGD and biologically plausible algorithm baselines.} Bold denotes the best performing algorithm, and italics denote the best runner-up algorithm. We see SGD and SAM are comparable, while significantly outperforming existing biologically plausible learning algorithms. For each experiment, we conduct five runs and calculate the error bars.}
  \renewcommand{\arraystretch}{1.2}
  \setlength{\tabcolsep}{3.5pt}
  \setlength{\belowrulesep}{-0.5pt}
  \setlength{\aboverulesep}{-0.5pt}
    \resizebox{\linewidth}{!}{%
    \begin{tabular}{c|ccccccc}
    \toprule
    Acc. $\uparrow$ & CIFAR10 & MNIST & Fashion MNIST & ChestMNIST & PathMNIST   & SVHN & STL10 \\
    \toprule
    \hline
    SGD & \textit{50.36 $\pm$ 0.16} & 98.44 $\pm$ 0.09  & \textit{88.63 $\pm$ 0.26} &  \textbf{73.26 $\pm$ 0.14}  & \textit{70.11 $\pm$ 0.10}  &  \textit{78.20 $\pm$ 0.13}  &  \textit{41.00 $\pm$ 0.07} \\
    
    FA & 47.85 $\pm$ 0.20 & 97.80 $\pm$ 0.19 & 85.41 $\pm$ 0.26 &  70.44 $\pm$ 0.27 & 69.90 $\pm$ 0.10 &  78.03 $\pm$ 0.15 & 39.29 $\pm$ 0.23 \\
    WM & 49.35 $\pm$ 0.13 & 98.40 $\pm$ 0.23 & 88.34 $\pm$ 0.36 &   70.14 $\pm$ 0.08 & 70.00 $\pm$ 0.15 &   78.08 $\pm$ 0.15 & 39.48 $\pm$ 0.19 \\
    KP Algorithm & 48.16 $\pm$ 0.09 & \textit{98.45 $\pm$ 0.09} & 88.52 $\pm$ 0.25 &  71.03 $\pm$ 0.10 &  69.51 $\pm$ 0.05 & \textit{78.20 $\pm$ 0.05} &  40.32 $\pm$ 0.11 \\
    \hline
    {SAM (ours)}  & \textbf{51.91 $\pm$ 0.21}  & \textbf{98.85 $\pm$ 0.10}  & \textbf{88.86 $\pm$ 0.17}  &  \textit{73.21 $\pm$ 0.11}   &  \textbf{70.23 $\pm$ 0.15}  &    \textbf{78.50 $\pm$ 0.07}   & \textbf{41.50 $\pm$ 0.10} \\
    \bottomrule
    \end{tabular}}%
  \label{tab:main}%
\end{table*}%

\subsection{SAM works efficiently in learning}

We evaluate our algorithm on seven widely used image classification datasets: CIFAR10 (\cite{krizhevsky2009}), MNIST (\cite{yann1998}), Fashion MNIST (\cite{xiao2017}), ChestMNIST (\cite{yang2021}, \cite{yang2023}), PathMNIST (\cite{yang2021}, \cite{yang2023}), SVHN (\cite{netzer2011}), and STL10 (\cite{coates2011}). For details about these datasets, see SI Experimental Setup. We compare our algorithm's performance against the standard backpropagation and other biologically plausible algorithms. These include Feedback Alignment (FA) (\cite{lillicrap2016}), which uses random fixed weights for error propagation instead of transposed weights; Weight Mirroring (WM) (\cite{akrout2019a}), which learns transposed feedback weights from noise in a separate phase; and the KP algorithm (\cite{akrout2019a, kolen1994}), which gives forward and feedback weights the same local update and weight decay, so they become transposes of each other during training. While the primary goal of this paper is not to pursue State-of-the-Art (SOTA) performance but to design a neural network algorithm that can be deployed in the brain, our approach achieves SOTA among biologically plausible algorithms on all datasets (see Table 1). Notably, our algorithm outperforms models trained with backpropagation on six out of the seven datasets. The only exception is ChestMNIST, where it performs 0.05\% lower than backpropagation.

\subsection{Overparameterization in the Downstream Pathway is Beneficial}
\label{over_downstream}
In the nervous system, the number of backprojections can be much larger than that of the forward stream. For example, the feedback connections V1 to LGN can be roughly 10 times the RGC synapses from the retina (\cite{briggs2020}). We have observed a potential computational explanation of this phenomenon in our simulations. Our models perform better as the feedback pathway becomes overparametrized and achieves the best performance across multiple datasets when the parameter count of the downstream pathway is approximately 5 to 8 times that of the upstream pathway. Specifically, as shown in Figure~3c for CIFAR10, performance peaks when the parameter count of the downstream pathway is 8 times that of the upstream pathway. We perform experiments on the three most challenging datasets from the seven datasets used in Table 1: CIFAR10 (\cite{krizhevsky2009}), SVHN (\cite{netzer2011}), and STL10 (\cite{coates2011}). 

\subsection{Learned Representations are Structurally Low-Rank}

Now, we show some examples of the learned weight matrices. We train a width-60 network on SVHN dataset for 50 epochs. See Figure~3a. As is common in deep learning, we plot the matrix $WW^T$ and also $\bar{V}\bar{V}^T$. We see that before training, these matrices are primarily diagonal matrices due to i.i.d. Gaussian initialization. After training, correlations between neurons and weights appear. We characterize the rank of a matrix through a continuous metric referred to as the effective rank (\cite{roy2007}); the right panel of Figure~3a shows that these weight matrices also become effectively low-rank across layers.

Figure~3a can be compared with similar matrices trained with SGD in \cite{ziyin2024} and \cite{huh2021}, for example, and they seem to have similar high-level structures, suggesting that our algorithm can lead to a meaningful representation learning. Detailed analysis of the learned representation using our algorithm may be an interesting future problem.

\subsection{Interconnection Plasticity is Crucial}

A remaining question is whether learning at some of the interconnections is more important than at others, and if so, which is more important? Additionally, are any interconnections not important at all? We perform this ablation study on CIFAR-10, where we initialize all the parameters randomly but only update a subset of all the connections. See Figure~3b. Note that $W$ is always plastic. To gauge how much the network has learned about the nonlinear relations in the data, we also compare with the learning trajectory of a linear model trained with SGD.

We discover the following phenomena:
\begin{itemize}[noitemsep,topsep=0pt, parsep=0pt,partopsep=0pt, leftmargin=13pt]
    \item Only updating $W$ works but only works as well as the linear model; making other parts plastic is thus important for learning nonlinear functions;
    \item If only one additional connection is made plastic, $\bar{V}$ is the most important one, improving performance from $40\%$ to roughly $46\%$;
    \item Making $\bar{V}$ plastic is similar to making $W,\ \bar{W},\ V$ plastic together, suggesting that some loss of learning capabilities due to connection lesions can be compensated by other connections (but not all);
    \item Making all four connections plastic improves the performance further, to the SGD level (cf. Table 1); this means that all connections need to be plastic to achieve the best performance.
\end{itemize}
Therefore, none of the pathways we introduced are redundant, and this partially explains the outperformance of our proposed algorithm over the existing two-pathway biologically plausible learning algorithms. Another reflection of the fact that there is an asymmetry between different pathways is that different connections require different learning time constants to work. In particular, the $\bar{V}$ matrices require the largest learning rates, usually four times that of $W$ (see SI Experimental Setup).

\begin{table*}[t!]
    \caption{\small \textbf{The upstream-downstream table of biologically plausible learning}, emphasizing the necessity for four types of connections between the two pathways: downstream to downstream ($d\to d$), downstream to upstream ($d\to u$), upstream to downstream ($u\to d$), and upstream to upstream ($u\to u$). Each connection must exhibit plasticity. For a learning rule to be biologically plausible, it must demonstrate both connectivity and plasticity.}
    \label{tab:table of plausibility}
    \centering
    \footnotesize
    \begin{tabular}{c|cccc|cccc}
    \hline
         & \multicolumn{4}{c|}{Connectivity}   & \multicolumn{4}{c}{Plasticity} \\
         & $u\to u$ & $d\to d$ & $u\to d$ & $d\to u$ & $u\to u$ & $d\to d$ & $u\to d$ & $d\to u$ \\
         \hline
     SGD & \cmark & \xmark & \xmark & \xmark & \cmark & \xmark & \xmark & \xmark \\
    Feedback Alignment & \cmark & \cmark & \xmark & \xmark & \cmark & \xmark & \xmark & \xmark \\
    Weight Mirroring & \cmark & \cmark & \xmark & \xmark & \cmark & \cmark & \xmark & \xmark \\
    SAM & \cmark & \cmark & \cmark & \cmark & \cmark & \cmark & \cmark & \cmark \\
    \hline
    \end{tabular}
\end{table*}

\subsection{How to identify these circuits}
\label{sec:how_to_identify}

A key question is whether and how one can identify the circuit we proposed in biological systems, especially how one can experimentally distinguish it from other algorithms of the feedback alignment type. In essence, this broad family of algorithms has four pathways (uu, ud, dd, du): (1) u (upstream) $\to$ u, (2) u (upstream) $\to$ d (down stream), (3) d $\to$ d, (4) d $\to$ u. The difference between these feedback type algorithms is in their prediction of the (a) anatomical structure and (b) plasticity of each of the pathways. See Table~\ref{tab:table of plausibility}. For example, a biological circuit is compatible with Feedback alignment (1) if uu and dd are randomly connected, and ud and du are not (which could happen, for example, in the cerebellum) and if (2) only uu is plastic, while dd, ud and du are not. In contrast, a biological circuit where all pathways are random and plastic will be compatible with our proposal, and not with the feedback alignment. Therefore, our proposal implies an eight-category classification of plausible learning algorithms, and biologists can experimentally test all eight criteria to decide which category applies to their system. 

There should not be an a priori belief that biological brains only use one such algorithm throughout the brain, and it is fully possible that different parts of the brain use different types of connections. For example, the feedback alignment motif seems especially suitable for the Purkinje cells because the climbing fibers are one-to-one, but it is very unlikely to be the right motif for, say, the cortices. For example, the apical dendrites of the L5 neurons, receiving largely feedback inputs, are found to be very plastic during motor learning (\cite{fu2011experience}). Similarly, the axon boutons of L6 neurons are highly plastic, and these axons often attach onto the basal dendrites of the L5 neurons, which could provide a feedback signal (\cite{fu2011experience}). There is, in any case, substantial evidence that spines in layer~1 (where many feedback inputs make their synapses) on both L5 and L2/3 pyramidal neurons are reasonably plastic across the adult mouse cortex, and that they show enhanced plasticity during learning paradigms (\cite{tattikota2017distinct}; \cite{yang2009stably}; \cite{xu2009rapid}; \cite{holtmaat2005transient}; \cite{makino2015learning}; \cite{larkum2013cellular}; \cite{abs2018learning}; \cite{doron2020perirhinal}); at these same feedback-targeted compartments, \cite{francioni2026vectorized} recently reported vectorized instructive signals at L5 apical dendrites during learning.



\subsection{An Example Implementation of SAM in Cortex}

Self-Assembling Motif (SAM) is defined in terms of four abstract pathway types—upstream-to-upstream (u$\to$u), upstream-to-downstream (u$\to$d), downstream-to-downstream (d$\to$d), and downstream-to-upstream (d$\to$u). These pathway types are functional constructs rather than claims about specific neuron morphologies or laminar structures. The defining architectural principles of SAM are (i) the existence of two interacting streams, (ii) driving interactions within each stream and modulatory interactions between streams, and (iii) the absence of closed loops composed entirely of driving connections within either stream. SAM itself makes no assumptions about the biophysical mechanisms by which these properties are implemented. When viewed through this functional lens, cortical lamination and its canonical connectivity patterns provide a useful and concrete example of how the SAM architecture could be realized in a biological system.

In sensory cortex, many pathways that operate within the same functional stream—such as those carrying representational content—are strong and driving, whereas interactions between distinct streams often exert a modulatory influence on activity and learning. Importantly, the distinction emphasized by SAM is not between feedforward and feedback pathways per se, but between within-stream and cross-stream interactions. Under this interpretation, a number of well-established cortical pathways can be organized into the four SAM categories. Table~\ref{tab:SAM_layer_correspondence} summarizes one such correspondence between canonical cortical pathways and SAM pathway types, together with a hypothesized relationship between dendritic targeting and the resulting driving or modulatory effects observed in cortex. Francioni et al.~\cite{francioni2026vectorized} report vectorized instructive signals at L5 apical dendrites during learning; this is consistent with the apical/distal targeting of the $d\to u$ pathway in the table.

\vspace{0.5em}

\noindent
\textbf{Interpretation and scope.}
The correspondence shown in Table~\ref{tab:SAM_layer_correspondence} should be interpreted as an example of how SAM could be implemented in cortex, rather than as a claim that SAM must be implemented through cortical lamination or that cortical lamination evolved specifically to implement SAM. Cortical areas exhibit substantial variation in laminar structure, connectivity, and cellular properties, and exceptions to canonical patterns are well documented. In particular, some cortical neurons, such as large layer~5 pyramidal neurons, exhibit conditional or gated driving as a result of their dendritic biophysics. Such properties reflect details of a specific biological implementation and are not assumptions or requirements of the SAM framework itself. Other brain systems may satisfy the same SAM constraints using different cellular or circuit mechanisms.

From the perspective of SAM, the central claim is therefore functional rather than morphological: learning systems that avoid unstable dynamics and signal contamination should exhibit two interacting streams, driving connectivity within each stream, modulatory connectivity between streams, and no closed loops of purely driving interactions within a stream. Cortical lamination provides one biologically plausible realization of these constraints, but not their only possible instantiation.

\begin{table*}[t!]
\centering
\caption{One possible correspondence between canonical cortical pathways and SAM pathway types. 
Dendritic targeting and observed influence type describe \emph{biological properties of the cortical implementation} rather than assumptions of the SAL/SAM framework. 
``Conditionally driving'' refers to a biophysical property observed in some layer~5 pyramidal neurons, arising from interactions between basal and apical compartments, and is \emph{not} a requirement of SAM itself.}
\label{tab:SAM_layer_correspondence}

\newcommand{\specialcell}[2][c]{%
  \begin{tabular}[#1]{@{}c@{}}#2\end{tabular}}

\newcommand{\specialcellL}[2][c]{%
  \begin{tabular}[#1]{@{}l@{}}#2\end{tabular}}

\begin{tabular}{lllll}
\hline
\specialcell{SAM \\ pathway} &
\specialcell{Weight \\ type} &
Cortical pathway example &
Dendritic targeting &
\specialcell{Observed \\ influence type} \\
\hline
u $\to$ u & $W$ &
L4 $\to$ L2/3 (within area) &
Basal / proximal &
Driving \\

 &  &
L2/3 $\to$ L4 (higher area) &
Basal / proximal &
Driving \\
\hline
u $\to$ d & $V$ &
L2/3 $\to$ L5 (within area) &
Basal $\,+$ apical &
\specialcellL{Conditionally \\ driving} \\
\hline
d $\to$ d & $\bar{W}$ &
L5 $\to$ (thalamus $\to$) L5 (inter-areal) &
Basal $\,+$ apical &
\specialcellL{Conditionally \\driving} \\

 &  &
L6 $\to$ (thalamus $\to$) L6 (inter-areal) &
Basal / proximal &
Driving \\
\hline
d $\to$ u & $\bar{V}$ &
L5 $\to$ L1 $\to$ L2/3 (inter-areal) &
Apical / distal &
Modulatory \\

 &  &
L5 $\to$ L2/3 (within area) &
Apical-biased &
Modulatory \\
\hline
\end{tabular}
\end{table*}

\subsection{Self-Organization Beyond the Two-Stream Scaffold}
 
The core of SAM is its local heterosynaptic learning rule; the two-stream motif
of the main text is only one setting in which we exercise it. In companion work
\cite{ziyin2025} we apply the same rule in a far more general setting---a
generalized SAM. A pool of neurons is randomly initialized and recurrently
connected, with no two-stream organization at all, and an evolutionary process
grows and prunes the connections. Driven by the same heterosynaptic plasticity,
the connectivity self-organizes into circuits that learn well, with the evolved
updates becoming aligned with the true gradient (see SI and \cite{ziyin2025}). This indicates that the heterosynaptic rule alone is enough to make a population of neurons self-organize into an effective learner, and that the SAL goal is achievable far beyond any specific scaffold---direct support for the broader SAL program.
  
\section{Discussion}

The theoretical and simulation results we present here demonstrate a new circuit architecture that not only emulates the computational abilities of conventional deep learning models but does so through mechanisms that align more closely with biological circuits. Specifically, our simulations show that a network trained with our algorithm achieves competitive performance on tasks such as image classification on CIFAR-10, closely matching or even surpassing traditional backpropagation systems in some cases. Theoretically, SAM implements generalized SGD when two explicit alignment conditions hold: interstream alignment $E_v^\ell\to 0$ (proved unconditionally when $\eta_V=\eta_{\bar V}$) and pathway alignment with $\|E_p^\ell\|$ small---in successful runs typically observed to decrease, and empirically the dominant source of gradient-approximation error once $E_v$ has tightened. Our proposed circuit does not pose any requirement on specificity of the connections. In fact, a surprising property of the architecture we propose is that the basic circuit self-assembles, emerging from random synaptic connections between the ascending and the descending stream. Intriguingly, the best computational performance is obtained when the number of backprojections is around 5 times the connections in the forward stream.

Optimization algorithms that can adjust the inner weights of a deep network are critical for state-of-the-art machine learning. This does not necessarily imply, however, that the same statement holds true for the brain. We have introduced a biologically plausible learning architecture that self-assembles from random initial connectivity and achieves SGD-like performance through local plasticity rules. Self-assembly requires an ascending and a descending stream – known to exist in all cortices – and initial random connections between them with synapses obeying reasonable local plasticity rules. It is also possible to use other biologically plausible versions of the regularization term we introduced, such as a Winner-Takes-All form of regularization. Such a mechanism can lead to one-to-one synapses between ascending and descending streams. It remains an open problem whether the cortex uses this mechanism. A brief discussion of the biological plausibility of the associated plasticity rules is presented in SI Biological Evidence for Synaptic Plasticity Rules. It is also possible that this circuit motif can be made more flexible and can potentially enable a large variety of motifs to learn; companion work \cite{ziyin2025} already explores this in a scaffold-free setting (SI section \emph{Gradient Learning Without a Two-Stream Scaffold}).
 
\textbf{Vectorized instructive signals in cortex.} \cite{francioni2026vectorized} recently reported vectorized instructive signals, including task error, at L5 apical dendrites during learning, with semi-independent somatic and dendritic computation; perturbing apical signaling disrupts learning. We take this as compatible with our example cortical SAM implementation (Table~\ref{tab:SAM_layer_correspondence}), where inter-areal feedback targets apical/distal compartments ($d\to u$, L5$\to$L1, modulatory).

\textbf{Temporal implementation and eligibility traces.} SAM is formulated in discrete time, with upstream inference preceding downstream error propagation and synaptic update. This ordering isolates the self-assembly mechanism and makes its connection to gradient learning analyzable, but it should not be read as a complete real-time biophysical model. The local rule requires only that a transient trace of each synapse's forward pre- and postsynaptic activity remain available when the feedback signal arrives---an eligibility trace, rather than a sustained copy of the full forward state---while the cross-stream weight that delivers the feedback signal is standing circuitry. The timing requirements of such continuous-time, biologically plausible learning are analyzed in~\cite{bacvanski2025continuous,bacvanski2026heteroassoc,bacvanski2026ctkp}. A fully continuous-time treatment of SAM is left to future work.

\textbf{Inference and learning in the descending stream.} The present implementation uses SAM in an error-feedback setting, with input supplied at one end of the motif and target or error information at the other. This is a useful controlled instantiation, but it is not a defining commitment of SAL/SAM. In cortex, descending pathways are recurrent and multiplexed, contributing to inference, attention, contextual modulation, and the resolution of sensory ambiguity. This motivates the broader aim of SAL: not to hand-specify one pathway as an inference stream and another as a special-purpose error-routing stream, but to let functional roles arise from uniform local rules. SAM is a step in this direction, because its two streams are homologous and governed by the same local plasticity principles; the present asymmetry comes from how the motif is driven, not from an internally distinct backward learning rule.

This perspective resonates with the motivation behind biologically inspired approaches such as predictive coding, which likewise seek to move beyond a simple input–error dichotomy by introducing predictive pathways and tying them to learning. SAL/SAM, however, aims at something more general: the goal is not to reproduce a particular cortical phenomenon—error assignment, or prediction—but to let pathways and their functional roles emerge, rather than assigning them by design.

Whether cortical feedback carries predictions, errors, or multiplexed combinations of them, and how such signals are read by other pathways, remains an open and substantial problem. Our claim is not that the present model already captures inference-bearing feedback, but that SAM provides a symmetric substrate in which inputs and errors are simply two kinds of signal that the motif does not handle differently. This opens a path: one could drive SAM with other signals in future work—for instance, to learn bidirectional symmetric associations—allowing the streams to acquire distinct representational, inferential, or credit-assignment roles. In the strongest form of the SAL ideal, even the predefined streams would be relaxed, with pathways, roles, and functional distinctions all emerging through self-organization. Companion work \cite{ziyin2025} takes a first step in this direction: with only the locus of error entry fixed, the same heterosynaptic rule in a randomly connected recurrent pool evolves credit-assignment circuitry without predefined inference or error-delivery streams (see SI). Further extending self-organization to fully multiplexed, inference-bearing feedback beyond the two-stream scaffold remains for future work, but these directions point to an appealing unification that the SAL perspective brings into view.

\textbf{Why ``evolution did it'' is not enough for realizability.} A tempting objection is that any biologically plausible circuit is realizable because evolution could have produced it. But realizability asks whether a proposed path \emph{guarantees} the structure we observe: $E$ explains $X$ only if $E$ would actually produce $X$. ``Evolution produced it'' is equally compatible with the circuit we found, a different circuit, or none at all; because it fits every outcome, it singles out none, and so establishes no causal link to the particular circuit in front of us. This is the same failure as weight transport---which asserts a correspondence with no path to it---and as passing a plausibility checklist, which shows only the absence of violations, not the presence of a path. SAM is offered as an actual path: the credit-assignment circuit assembles from random connectivity under uniform local rules, so what others assume, it derives.

\textbf{Experimental predictions.} Our results indicate that biologically plausible learning algorithms can rival established machine learning algorithms while respecting the brain's biophysical constraints. Our work makes several predictions that are, at least in principle, experimentally verifiable: 1) Synaptic Motifs - Reciprocal synaptic connections between ascending and descending pathways with four plastic synapses ($W$, $\bar{W}$, $V$, $\bar{V}$) should be generally present in cortical networks; 2) Heterosynaptic Plasticity - Changes in synaptic strength should depend on the activity of other synapses (these synapses must be on the same neuron but do not need to be electrically close, depending on the specific mechanism of heterosynaptic plasticity); 3) Self-Assembly - The initially random connections between the two streams assemble over development/learning, guided by heterosynaptic plasticity rules; 4) Distinct Pathways - The back projections may exhibit different electrophysiological properties, such as nonlinearities depending on synaptic inputs, compared to feedforward pathways, also possibly with different time constants as the computational model is found to work the best when different connections have different learning time constants; 5) Relative size of the streams: Our results indicate that learning produces better performance when the feedback connections are more numerous than the feedforward. Experimental verification of these predictions may transform our understanding of learning in both biological and artificial systems.

\matmethods{Computational experiments were implemented in PyTorch on NVIDIA GPUs. The circuit architecture (upstream and downstream pathways with four synaptic connection matrices $W$, $\bar{W}$, $V$, $\bar{V}$) and heterosynaptic plasticity rules are given in the main text. Full experimental details (datasets, baselines, architectures, training schedules, hyperparameter selection, and hardware) are provided in SI Experimental Setup and SI Alignment Diagnostics. Proofs and algorithm pseudocode are provided in SI Theory and SI Pseudo Code. Further discussion is provided in SI Biological Evidence for Synaptic Plasticity Rules and SI Gradient Learning Without a Two-Stream Scaffold.}

\showmatmethods{}

\dataavail{All code and data are on GitHub at \url{https://github.com/liaoq/SAM2026}, including the \textbf{2024/2025} benchmark implementation (SI Algorithm~1) and the \textbf{2026} implementation used for theory and alignment diagnostics (SI Algorithm~2).}

\acknow{We thank the reviewers for their exceptionally constructive feedback. This work was supported by the K. Lisa Yang Integrative Computational Neuroscience (ICoN) Center, part of the Yang Tan Collective at MIT; the Center for Brains, Minds and Machines (CBMM, NSF STC award CCF-1231216); and the McGovern Institute for Brain Research at MIT. L.Z. acknowledges support from NTT Research. M.T.H. was additionally supported by the NIH (R01MH135141).
}

\showacknow{}

\bibliography{pnas_ref_corrected}


\clearpage
\titleformat{\paragraph}[runin]
  {\normalfont\normalsize\bfseries}
  {}
  {1em}
  {#1}




\maketitle

\SItext

\appendix

\section{Theory}
\label{sec:theory}

This section supplements the main text with the full ReLU-only alignment proof.
The argument is organized in three layers:
Lemma~\ref{lem:Ev} proves \emph{unconditional} interstream decay $E_v^\ell\to 0$ when $\eta_V=\eta_{\bar V}$;
Lemma~\ref{lem:depth-chain} telescopes the depth chain in the primitive errors $E_p,E_v$ (neither is assumed zero);
Theorem~\ref{thm:main} converts that chain into the SAL weight update with effective prefactor $H_D^\ell+\widetilde E_{v,D}^\ell$.

\subsection{Setup}\label{sec:notation}

\textbf{SAL} / \textbf{SAM}: dual-stream learning (main text).
Layers $0,\ldots,L$; readout at $L$.
Upstream: $h^{\ell+1}=W^\ell D_u^\ell h^\ell$ with $D_u^\ell=\diag(\mathbf{1}_{h^\ell>0})$; define the upstream activity port $a^\ell:=D_u^\ell h^\ell$.
Downstream pre-activation $\bar h^\ell$ and post-activation $\bar a^\ell:=D_d^\ell\bar h^\ell$ obey
\begin{equation}\label{eq:hbar-rec}
    \bar h^\ell = \bar W^\ell\bar a^{\ell+1},
    \qquad
    \bar a^L = -g^L.
\end{equation}
Interstream maps $V^\ell,\bar V^\ell$; $E_v^\ell:=V^\ell-(\bar V^{\ell-1})^T$.
Cotangent at depth $k$:
\begin{equation}\label{eq:g-def}
    g^k := \frac{\partial F}{\partial h^k} \in\R^{d_k}.
\end{equation}

Composed maps from readout to motif $\ell$ (superscript $\ell$ = ``from $L$ down to $\ell$''):
\begin{align}
    J_{\rm SGD}^{\ell} &:= T^{\ell+1}\cdots T^{L-1},
    \label{eq:Jsgd}\\
    \bar J_{\rm SAM}^{\ell} &:= \bar S^{\ell+1}\cdots \bar S^{L-1}
    = \textstyle\prod_{k=\ell+1}^{L-1}D_d^k\bar W^k,
    \label{eq:barJsam}\\
    J_{\rm SAM}^{\ell} &:= \bar V^\ell\,\bar J_{\rm SAM}^{\ell}.
    \label{eq:Jsam}
\end{align}
Exact autograd: $g^{\ell+1}=J_{\rm SGD}^{\ell}g^L$.
Exact SAM: $\bar a^{\ell+1}=\bar J_{\rm SAM}^{\ell}\bar a^L$.

\subsection{One-step maps and telescoping target}\label{sec:one-step}

Maps are chosen so depth composition telescopes to $\bar J_{\rm SAM}^{\ell}$:
\begin{equation}\label{eq:telescope-goal}
    J_{\rm SGD}^{\ell}
    =
    T^{\ell+1}\cdots T^{L-1}
    \;\stackrel{?}{=}\;
    (V^{\ell+1})^+\bar J_{\rm SAM}^{\ell},
    \qquad
    \bar J_{\rm SAM}^{\ell}=\bar S^{\ell+1}\cdots \bar S^{L-1}.
\end{equation}
Interior join in $\prod S^k$ carries the splice (write $P_k:=V^k(V^k)^+$):
\begin{equation}\label{eq:splice-I}
    (\bar V^{k-1})^T(V^k)^+ = P_k - E_v^k(V^k)^+,
    \qquad
    E_v^k:=V^k-(\bar V^{k-1})^T.
\end{equation}

\paragraph{Upstream cotangent port ($g^{\ell+1}$).}
\begin{equation}\label{eq:T-def}
    T^\ell := D_u^\ell (W^\ell)^T,
    \qquad
    \mathcal{B}^\ell := D_d^\ell\bar W^\ell(\bar V^\ell)^T,
    \qquad
    S^\ell := (V^\ell)^+\mathcal{B}^\ell,
    \qquad
    \mathcal{F}^\ell := V^\ell T^\ell.
\end{equation}

\paragraph{Downstream activity port ($\bar a^{\ell+1}$).}
\begin{equation}\label{eq:barS-def}
    \bar S^\ell := D_d^\ell\bar W^\ell.
\end{equation}

\subsection{Main results}\label{sec:main-result}

\begin{definition}[Primitive alignment errors]\label{def:primitive-errors}
\begin{align}
    E_v^\ell &:= V^\ell-(\bar V^{\ell-1})^T,
    \label{eq:Ev-def}\\
    E_p^\ell &:= \mathcal{F}^\ell-\mathcal{B}^\ell
    = V^\ell D_u^\ell (W^\ell)^T - D_d^\ell\bar W^\ell(\bar V^\ell)^T.
    \label{eq:Ep-def}
\end{align}
\end{definition}

\begin{lemma}[Unconditional $E_v$ decay]\label{lem:Ev}
Fix motif $\ell$ and sample $x$.
Recall $E_v^\ell:=V^\ell-(\bar V^{\ell-1})^T$ \eqref{eq:Ev-def}.
Under the SAL interstream updates with matched rates $\eta_V=\eta_{\bar V}$, every training step satisfies
\begin{equation}\label{eq:Ev-dynamics}
    \Delta E_v^\ell = -\gamma E_v^\ell,
\end{equation}
hence $E_v^\ell(t)=(1-\gamma)^t E_v^\ell(0)$, and in the continuous-time limit $E_v^\ell\to 0$ exponentially for every $\gamma>0$.
Equivalently, at $E_v^\ell=0$ one has $\bar V^{\ell-1}=(V^\ell)^T$; this does \textbf{not} require $\Delta V^\ell=0$ or $\Delta\bar V^{\ell-1}=0$.

\smallskip
\noindent\textbf{Scope.}
The conclusion is independent of pathway alignment ($E_p^\ell$), the loss value, mini-batch noise, and depth-chain residuals.
It uses only the outer-product structure of the interstream updates and matched learning rates.
Neither $E_p^\ell\to 0$ nor small $\|\Delta V^\ell\|$ is assumed or implied.
\end{lemma}

\begin{proof}
The SAL interstream plasticity rules are
\begin{align}
    \Delta V^\ell
    &= \eta_V\,\bar a^\ell(D_u^\ell h^\ell)^T - \gamma V^\ell,
    \label{eq:dV}\\
    \Delta \bar V^\ell
    &= \eta_{\bar V}\,(D_u^{\ell+1}h^{\ell+1})(\bar a^{\ell+1})^T - \gamma \bar V^\ell,
    \label{eq:dbarV}
\end{align}
with the same local decay $\gamma>0$ on both streams.
Transpose \eqref{eq:dbarV} at layer $\ell{-}1$:
\[
    \Delta(\bar V^{\ell-1})^T
    =
    \eta_{\bar V}\,\bar a^\ell(D_u^\ell h^\ell)^T - \gamma(\bar V^{\ell-1})^T.
\]
Subtract from \eqref{eq:dV}:
\begin{align*}
    \Delta E_v^\ell
    &= \Delta V^\ell - \Delta(\bar V^{\ell-1})^T \\
    &= \bigl(\eta_V-\eta_{\bar V}\bigr)\bar a^\ell(D_u^\ell h^\ell)^T
       - \gamma\bigl(V^\ell-(\bar V^{\ell-1})^T\bigr).
\end{align*}
When $\eta_V=\eta_{\bar V}$, the shared Hebbian term $\bar a^\ell(D_u^\ell h^\ell)^T$ cancels and \eqref{eq:Ev-dynamics} follows from $E_v^\ell=V^\ell-(\bar V^{\ell-1})^T$.
\end{proof}

\begin{assumption}[Standing hypotheses]\label{ass:standing}
$\rank(V^\ell)=d_\ell$ (so $(V^\ell)^+V^\ell=I$); $\eta_V=\eta_{\bar V}$;
$(\bar V^{L-1})^T=I$ and $\bar a^L=-g^L$.
\end{assumption}

\begin{lemma}[Depth chain]\label{lem:depth-chain}
Fix sample $x$ and motif $\ell\in\{0,\ldots,L-2\}$.
Suppose Assumption~\ref{ass:standing}, $\|E_p^k(x)\|\le\epsilon_p^k(x)$ for $k=\ell+1,\ldots,L-1$, and $\|E_v^k\|\le\epsilon_v^k$ for $k=\ell+2,\ldots,L-1$.
Then
\begin{equation}\label{eq:depth-chain}
    J_{\rm SGD}^{\ell}
    =
    (V^{\ell+1})^+\bar J_{\rm SAM}^{\ell}
    + R_{\rm chain}^{\ell},
\end{equation}
where $R_{\rm chain}^{\ell}=R_T^\ell+R_S^\ell$ and the two exact transported-error bounds are given in \eqref{eq:I4-RT-bound} and \eqref{eq:I5-RS-bound}.
In particular, $\|R_{\rm chain}^{\ell}\|\le\|R_T^{\ell}\|+\|R_S^{\ell}\|$ is linear in the displayed primitive-error bounds.
Exact equality $R_{\rm chain}^{\ell}=0$ holds when $E_p^k=E_v^k=0$ at all relevant depths.
\end{lemma}

\begin{theorem}[SAL $\approx$ generalized SGD]\label{thm:main}
Fix sample $x$ and motif $\ell\in\{0,\ldots,L-2\}$.
Suppose the hypotheses of Lemma~\ref{lem:depth-chain}.
Suppose additionally that $\|E_v^{\ell+1}\|\le\epsilon_v^{\ell+1}$.
Let $H^\ell:=\bar V^\ell(\bar V^\ell)^T$ and $\widetilde E_v^\ell:=\bar V^\ell E_v^{\ell+1}$ (identified in Step~I6 below), and define the effective weight-update operators
\[
    H_D^\ell:=D_u^{\ell+1}H^\ell D_u^{\ell+1},
    \qquad
    \widetilde E_{v,D}^\ell
    :=D_u^{\ell+1}\widetilde E_v^\ell D_u^{\ell+1}.
\]
Then the SAL rule satisfies
\begin{equation}\label{eq:main}
    \Delta_{\rm SAL}W^\ell
    =
    \eta_W\,\bigl(H_D^\ell+\widetilde E_{v,D}^\ell\bigr)\,
    \Delta_{\rm sgd}W^\ell
    -\gamma W^\ell
    + R^\ell,
\end{equation}
with $\Delta_{\rm sgd}W^\ell=-\nabla_{W^\ell}F$.
The residual is
\[
    R^\ell
    :=
    -\eta_W D_u^{\ell+1}
    (\mathcal R_{\rm Jac}^\ell g^L)(D_u^\ell h^\ell)^T,
\]
and obeys
\begin{equation}\label{eq:R-bound}
    \|R^\ell\|
    \le
    \eta_W\|D_u^{\ell+1}\|\,\|g^L\|\,\|D_u^\ell h^\ell\|
    \left(
        \|\bar V^\ell V^{\ell+1}\|\,\|R_{\rm chain}^\ell\|
        +\epsilon_v^{\ell+1}\|\bar J_{\rm SAM}^\ell\|
    \right),
\end{equation}
where $\|R_{\rm chain}^\ell\|$ is bounded by $\|R_T^\ell\|+\|R_S^\ell\|$ via \eqref{eq:I4-RT-bound} and \eqref{eq:I5-RS-bound}.
Also,
\[
    \|\widetilde E_{v,D}^\ell\|
    \le \|\bar V^\ell\|\,\epsilon_v^{\ell+1}.
\]
Here $\|D_u^{\ell+1}\|\le1$ for a ReLU gate.
When $E_p^k=E_v^k=0$ at all relevant depths, $R^\ell=0$ and
\[
    \Delta_{\rm SAL}W^\ell
    =
    \eta_W H_D^\ell\Delta_{\rm sgd}W^\ell
    -\gamma W^\ell.
\]
The aligned preconditioner is positive semidefinite because
\[
    H_D^\ell
    =
    (D_u^{\ell+1}\bar V^\ell)(D_u^{\ell+1}\bar V^\ell)^T
    \succeq0.
\]
If $\bar V^\ell$ remains bounded, Lemma~\ref{lem:Ev} implies
$\widetilde E_{v,D}^\ell\to0$, so the finite-error operator in \eqref{eq:main} approaches this PSD matrix.
\end{theorem}

\begin{remark}[What is proved vs.\ conditional]\label{rem:early-Ev}
Lemma~\ref{lem:Ev}: $E_v^\ell$ decays unconditionally.
Lemma~\ref{lem:depth-chain}: depth telescoping in $E_p,E_v$ (neither assumed zero).
Theorem~\ref{thm:main}: SAL weight update from that chain.
Early training with large $\epsilon_v^k$ inflates $\|R_{\rm chain}^{\ell}\|$ through the transported interior-join terms in $R_S^\ell$ before $E_v$ decays.
Decrease of $E_p^\ell$ from the learning rules alone is not proved here, but in experiments $\|E_p^\ell\|$ and $\|E_p^\ell\|/\|\mathcal{F}^\ell\|$ usually fall during stable training and can rise when a run becomes unstable; when and why $E_p^\ell$ improves or worsens is left to future work.
\end{remark}

\subsection{Proof of Lemma~\ref{lem:depth-chain} and Theorem~\ref{thm:main}}\label{sec:proof-main}

Assume $\|E_p^k\|\le\epsilon_p^k$, $\|E_v^k\|\le\epsilon_v^k$, and Assumption~\ref{ass:standing}.
Throughout, fix a sample $x$ and omit explicit dependence on~$x$.
Write $P_k:=V^k(V^k)^+$.

\paragraph{I1 (one-layer bridge).}
By Definition~\ref{def:primitive-errors}, $\mathcal{F}^k=\mathcal{B}^k+E_p^k$.
Left-multiply by $(V^k)^+$:
\begin{equation}\label{eq:I1-T-S}
    T^k = S^k + (V^k)^+E_p^k,
    \qquad
    S^k=(V^k)^+\mathcal{B}^k,
    \qquad
    \|(V^k)^+E_p^k\|\le \|(V^k)^+\|\,\epsilon_p^k.
\end{equation}

\paragraph{I2 (splice).}
The interstream splice is \eqref{eq:splice-I}:
\[
    (\bar V^{k-1})^T(V^k)^+ = P_k - E_v^k(V^k)^+.
\]
When $E_v^k=0$ this is the orthogonal projector onto $\col(V^k)$.

\paragraph{I3 (projector on the pathway).}
Since $\mathcal{F}^k=V^k T^k\in\col(V^k)$, $P_k\mathcal{F}^k=\mathcal{F}^k$. Write $\mathcal{B}^k=\mathcal{F}^k-E_p^k$:
\begin{equation}\label{eq:I3-PB-expand}
    P_k\mathcal{B}^k
    =
    P_k(\mathcal{F}^k-E_p^k)
    =
    \mathcal{F}^k - P_k E_p^k
    =
    \mathcal{B}^k + (I-P_k)E_p^k,
\end{equation}
or equivalently $(I-P_k)\mathcal{B}^k=-(I-P_k)E_p^k$.

\paragraph{I4 (product-difference lemma; $T\to S$).}
Start from \eqref{eq:Jsgd} and substitute the one-layer split \eqref{eq:I1-T-S} at every depth:
\begin{equation}\label{eq:I4-T-expand}
    J_{\rm SGD}^{\ell}
    =
    T^{\ell+1}\cdots T^{L-1}
    =
    \prod_{k=\ell+1}^{L-1}\bigl(S^k+(V^k)^+E_p^k\bigr).
\end{equation}
Empty products in this and subsequent displays are identities.
The two products are compared by the following product-difference lemma, which carries exactly one slot difference $(A_k-B_k)$ per summand.

\begin{lemma}[Product-difference lemma]\label{lem:prod-diff}
For conformable matrix sequences $A_k,B_k$, $k=m,\ldots,n$,
\begin{equation}\label{eq:prod-diff}
    \prod_{k=m}^{n}A_k-\prod_{k=m}^{n}B_k
    =
    \sum_{k=m}^{n}
    \Bigl(\prod_{j=m}^{k-1}A_j\Bigr)
    \bigl(A_k-B_k\bigr)
    \Bigl(\prod_{j=k+1}^{n}B_j\Bigr).
\end{equation}
\end{lemma}

\begin{proof}
Define the hybrid products
$X_j:=A_m\cdots A_j\,B_{j+1}\cdots B_n$ for $j=m-1,\ldots,n$, so that
$X_{m-1}=\prod_{k=m}^nB_k$ and $X_n=\prod_{k=m}^nA_k$.
For each $j=m,\ldots,n$ the two hybrids $X_j$ and $X_{j-1}$ differ in the single slot~$j$, whence
\[
    X_j-X_{j-1}
    =
    A_m\cdots A_{j-1}\,(A_j-B_j)\,B_{j+1}\cdots B_n .
\]
Summing over $j=m,\ldots,n$ telescopes the left-hand side to $X_n-X_{m-1}$, which is \eqref{eq:prod-diff}.
\end{proof}

Apply Lemma~\ref{lem:prod-diff} with $A_k=T^k$, $B_k=S^k$ over $k=\ell+1,\ldots,L-1$; by I1 \eqref{eq:I1-T-S} the slot differences are exactly $T^k-S^k=(V^k)^+E_p^k$.
This gives
\begin{equation}\label{eq:I4-S-plus-RT}
    J_{\rm SGD}^{\ell}
    =
    S^{\ell+1}\cdots S^{L-1} + R_T^{\ell},
    \qquad
    R_T^\ell
    :=
    \sum_{k=\ell+1}^{L-1}
    T^{\ell+1}\cdots T^{k-1}
    (V^k)^+E_p^k
    S^{k+1}\cdots S^{L-1}.
\end{equation}
Each summand of $R_T^\ell$ is linear in a single $E_p^k$.
By submultiplicativity applied to each summand,
\begin{equation}\label{eq:I4-RT-bound}
    \|R_T^{\ell}\|
    \le
    \sum_{k=\ell+1}^{L-1}
    \Bigl(\prod_{j=\ell+1}^{k-1}\|T^j\|\Bigr)
    \Bigl(\prod_{j=k+1}^{L-1}\|S^j\|\Bigr)
    \|(V^k)^+\|\,\epsilon_p^k .
\end{equation}
It remains to telescope the $S$-product to the SAM tail $(V^{\ell+1})^+\bar J_{\rm SAM}^{\ell}$ \eqref{eq:telescope-goal}.

\paragraph{I5 ($S$-product induction).}
Write $S^k=(V^k)^+\mathcal{B}^k$.
The single join identity behind the induction is
\begin{equation}\label{eq:I5-join}
    (\bar V^{k-1})^T(V^k)^+\mathcal{B}^k
    =
    P_k\mathcal{B}^k-E_v^k S^k
    =
    \mathcal{B}^k+\bigl[(I-P_k)E_p^k-E_v^k S^k\bigr],
\end{equation}
by the splice \eqref{eq:splice-I} followed by the projector expansion \eqref{eq:I3-PB-expand}.

\smallskip
\noindent\emph{Partial products.}
For $m=\ell+1,\ldots,L-1$ set
\begin{equation}\label{eq:I5-QA}
    Q_m:=S^{\ell+1}\cdots S^{m},
    \qquad
    A_m:=(V^{\ell+1})^+\bar S^{\ell+1}\cdots\bar S^{m},
\end{equation}
so that $Q_{m+1}=Q_mS^{m+1}$ and $A_{m+1}=A_m\bar S^{m+1}$.

\smallskip
\noindent\emph{Claim (induction on $m$).}
For every $m=\ell+1,\ldots,L-1$,
\begin{equation}\label{eq:I5-IH}
    Q_m=A_m(\bar V^m)^T+\rho_m,
    \qquad
    \rho_{\ell+1}:=0,
    \qquad
    \rho_{m+1}
    :=
    \rho_m S^{m+1}
    +A_m\bigl[(I-P_{m+1})E_p^{m+1}-E_v^{m+1}S^{m+1}\bigr].
\end{equation}

\noindent\emph{Base case $m=\ell+1$.}
By the definitions \eqref{eq:T-def} and \eqref{eq:barS-def},
\[
    Q_{\ell+1}=S^{\ell+1}=(V^{\ell+1})^+\mathcal{B}^{\ell+1}
    =(V^{\ell+1})^+\bar S^{\ell+1}(\bar V^{\ell+1})^T
    =A_{\ell+1}(\bar V^{\ell+1})^T,
\]
so \eqref{eq:I5-IH} holds with $\rho_{\ell+1}=0$.
The entry join $\ell+1$ therefore contributes no residual; this is why the interior-join sum below starts at $k=\ell+2$.

\noindent\emph{Inductive step $m\to m+1$ (for $\ell+1\le m\le L-2$).}
Assume \eqref{eq:I5-IH} at $m$. Then
\begin{align}
    Q_{m+1}
    &=Q_mS^{m+1}
    =A_m(\bar V^m)^T(V^{m+1})^+\mathcal{B}^{m+1}+\rho_mS^{m+1}
    \label{eq:I5-step1}\\
    &=A_m\bigl[\mathcal{B}^{m+1}+(I-P_{m+1})E_p^{m+1}-E_v^{m+1}S^{m+1}\bigr]
      +\rho_mS^{m+1}
    \label{eq:I5-step2}\\
    &=A_m\bar S^{m+1}(\bar V^{m+1})^T
      +\bigl[\rho_mS^{m+1}
             +A_m\bigl((I-P_{m+1})E_p^{m+1}-E_v^{m+1}S^{m+1}\bigr)\bigr]
    \label{eq:I5-step3}\\
    &=A_{m+1}(\bar V^{m+1})^T+\rho_{m+1},
    \label{eq:I5-step4}
\end{align}
using $S^{m+1}=(V^{m+1})^+\mathcal{B}^{m+1}$ in \eqref{eq:I5-step1}, the join identity \eqref{eq:I5-join} at $k=m+1$ in \eqref{eq:I5-step2}, and $\mathcal{B}^{m+1}=\bar S^{m+1}(\bar V^{m+1})^T$ in \eqref{eq:I5-step3}.

\noindent\emph{Unrolling and closing at the readout.}
Solving the recursion in \eqref{eq:I5-IH} from $\rho_{\ell+1}=0$ gives
\begin{equation}\label{eq:I5-rho-unrolled}
    \rho_{L-1}
    =
    \sum_{k=\ell+2}^{L-1}
    A_{k-1}
    \bigl[(I-P_k)E_p^k-E_v^k S^k\bigr]
    S^{k+1}\cdots S^{L-1}.
\end{equation}
At $m=L-1$, Assumption~\ref{ass:standing} gives $(\bar V^{L-1})^T=I$, and $A_{L-1}=(V^{\ell+1})^+\bar S^{\ell+1}\cdots\bar S^{L-1}=(V^{\ell+1})^+\bar J_{\rm SAM}^{\ell}$ by \eqref{eq:barJsam}.
Hence \eqref{eq:I5-IH} at $m=L-1$ reads
\begin{equation}\label{eq:I5-S-telescope}
    S^{\ell+1}\cdots S^{L-1}
    =
    (V^{\ell+1})^+\bar J_{\rm SAM}^{\ell} + R_S^{\ell},
\end{equation}
where
\begin{equation}\label{eq:I5-RS-exact}
    R_S^{\ell}
    :=\rho_{L-1}
    =
    \sum_{k=\ell+2}^{L-1}
    A_{k-1}
    \bigl[(I-P_k)E_p^k-E_v^k S^k\bigr]
    \bigl(S^{k+1}\cdots S^{L-1}\bigr),
\end{equation}
and $A_{k-1}=(V^{\ell+1})^+\bar S^{\ell+1}\cdots\bar S^{k-1}$.
When $E_p^k=E_v^k=0$ at every interior join, $R_S^\ell=0$ and $Q_{L-1}=A_{L-1}=(V^{\ell+1})^+\bar J_{\rm SAM}^{\ell}$.
By submultiplicativity applied to each summand of \eqref{eq:I5-RS-exact},
\begin{align*}
    &\bigl\|A_{k-1}\bigl[(I-P_k)E_p^k-E_v^k S^k\bigr]
      (S^{k+1}\cdots S^{L-1})\bigr\|\\
    &\qquad\le
    \|A_{k-1}\|
    \bigl(\epsilon_p^k+\epsilon_v^k\|S^k\|\bigr)
    \prod_{j=k+1}^{L-1}\|S^j\|,
\end{align*}
with $\|I-P_k\|\le1$.
The ideal accumulator obeys $\|A_{k-1}\|\le\|(V^{\ell+1})^+\|M^{k-\ell-1}$
for any $M\ge\max_j\|\bar S^j\|$.
Summing over interior joins,
\begin{equation}\label{eq:I5-RS-bound}
    \|R_S^{\ell}\|
    \le
    \|(V^{\ell+1})^+\|
    \sum_{k=\ell+2}^{L-1}
    M^{\,k-\ell-1}
    \bigl(\epsilon_p^k+\epsilon_v^k\|S^k\|\bigr)
    \prod_{j=k+1}^{L-1}\|S^j\|.
\end{equation}

\smallskip
\noindent\textbf{Combine $R_T^{\ell}$ and $R_S^{\ell}$.}
Substitute \eqref{eq:I5-S-telescope} into \eqref{eq:I4-S-plus-RT}:
\begin{equation}\label{eq:I5-depth}
    J_{\rm SGD}^{\ell}
    =
    (V^{\ell+1})^+\bar J_{\rm SAM}^{\ell}
    + R_S^{\ell}
    + R_T^{\ell},
    \qquad
    R_{\rm chain}^{\ell}:=R_S^{\ell}+R_T^{\ell}.
\end{equation}
Adding \eqref{eq:I4-RT-bound} and \eqref{eq:I5-RS-bound} gives $\|R_{\rm chain}^{\ell}\|\le\|R_T^{\ell}\|+\|R_S^{\ell}\|$, which is linear in the primitive errors $\epsilon_p$ and $\epsilon_v$.
If in addition every $\|T^j\|$ and $\|S^j\|$ is bounded by the same $M$, a coarse envelope is $O\bigl(M^{L-\ell-1}\sum(\epsilon_p+\epsilon_v)\bigr)$.

\paragraph{I6 (SAL update $\leftrightarrow$ generalized SGD).}
Start from I5 \eqref{eq:I5-depth}:
\begin{equation}\label{eq:I6-I5-repeat}
    J_{\rm SGD}^{\ell}
    =
    (V^{\ell+1})^+\bar J_{\rm SAM}^{\ell}
    + R_{\rm chain}^{\ell}.
\end{equation}

\textbf{I6.1 (left-multiply by $V^{\ell+1}$).}
Left-multiply \eqref{eq:I6-I5-repeat} by $V^{\ell+1}$ and set $P_{\ell+1}:=V^{\ell+1}(V^{\ell+1})^+$:
\begin{equation}\label{eq:I6-P-barJ}
    P_{\ell+1}\bar J_{\rm SAM}^{\ell}
    =
    V^{\ell+1}J_{\rm SGD}^{\ell}
    - V^{\ell+1}R_{\rm chain}^{\ell}.
\end{equation}

\textbf{I6.2 (keep the entry projector).}
SAL uses the full downstream Jacobian, so decompose it into its component in $\col(V^{\ell+1})$ and its orthogonal leakage:
\begin{equation}\label{eq:I6-projector-split}
    J_{\rm SAM}^{\ell}
    =
    \bar V^\ell P_{\ell+1}\bar J_{\rm SAM}^{\ell}
    +
    \bar V^\ell(I-P_{\ell+1})\bar J_{\rm SAM}^{\ell}.
\end{equation}
For the first term, use \eqref{eq:I6-P-barJ}.
For the second, $P_{\ell+1}V^{\ell+1}=V^{\ell+1}$ implies
\begin{equation}\label{eq:I6-VTP}
    (V^{\ell+1})^T(I-P_{\ell+1})=0 .
\end{equation}
Transposing $E_v^{\ell+1}=V^{\ell+1}-(\bar V^\ell)^T$ gives
$\bar V^\ell=(V^{\ell+1})^T-(E_v^{\ell+1})^T$, so by \eqref{eq:I6-VTP}
\begin{equation}\label{eq:I6-B1key}
    \bar V^\ell(I-P_{\ell+1})
    =
    -(E_v^{\ell+1})^T(I-P_{\ell+1}).
\end{equation}
Therefore
\begin{equation}\label{eq:I6-entry-leakage}
    \bar V^\ell(I-P_{\ell+1})\bar J_{\rm SAM}^{\ell}
    =
    -(E_v^{\ell+1})^T(I-P_{\ell+1})\bar J_{\rm SAM}^{\ell}.
\end{equation}

\textbf{I6.3 (assemble $J_{\rm SAM}$).}
Substituting \eqref{eq:I6-P-barJ} and \eqref{eq:I6-entry-leakage} into
\eqref{eq:I6-projector-split} gives the exact relation
\begin{equation}\label{eq:I6-Jsam-expanded}
    J_{\rm SAM}^{\ell}
    =
    \bar V^\ell V^{\ell+1} J_{\rm SGD}^{\ell}
    - \bar V^\ell V^{\ell+1} R_{\rm chain}^{\ell}
    - (E_v^{\ell+1})^T(I-P_{\ell+1})\bar J_{\rm SAM}^{\ell}.
\end{equation}

\textbf{I6.4 (split $H^\ell$ from $E_v^{\ell+1}$).}
Define (as in Theorem~\ref{thm:main})
\begin{equation}\label{eq:H-def}
    H^\ell:=\bar V^\ell(\bar V^\ell)^T,
    \qquad
    \widetilde E_v^\ell:=\bar V^\ell E_v^{\ell+1}.
\end{equation}
With $V^{\ell+1}=(\bar V^\ell)^T+E_v^{\ell+1}$ \eqref{eq:Ev-def},
\begin{equation}\label{eq:I6-V-split}
    \bar V^\ell V^{\ell+1}
    =
    H^\ell + \widetilde E_v^\ell.
\end{equation}
Substituting into \eqref{eq:I6-Jsam-expanded},
\begin{equation}\label{eq:I6-Jsam-H-split}
    J_{\rm SAM}^{\ell}
    =
    \bigl(H^\ell+\widetilde E_v^\ell\bigr) J_{\rm SGD}^{\ell}
    - \bar V^\ell V^{\ell+1} R_{\rm chain}^{\ell}
    - (E_v^{\ell+1})^T(I-P_{\ell+1})\bar J_{\rm SAM}^{\ell}.
\end{equation}

\textbf{I6.5 (Jacobian relation).}
Define
\begin{equation}\label{eq:RJac-def}
    \mathcal{R}_{\rm Jac}^{\ell}
    :=
    - \bar V^\ell V^{\ell+1} R_{\rm chain}^{\ell}
    - (E_v^{\ell+1})^T(I-P_{\ell+1})\bar J_{\rm SAM}^{\ell},
\end{equation}
so
\begin{equation}\label{eq:I6-Jac}
    J_{\rm SAM}^{\ell}
    =
    \bigl(H^\ell+\widetilde E_v^\ell\bigr) J_{\rm SGD}^{\ell}
    + \mathcal{R}_{\rm Jac}^{\ell}.
\end{equation}
When $R_{\rm chain}^{\ell}=0$ and $E_v^{\ell+1}=0$, this reduces to $J_{\rm SAM}^{\ell}=H^\ell J_{\rm SGD}^{\ell}$.
Moreover,
\begin{equation}\label{eq:RJac-bound}
    \|\mathcal{R}_{\rm Jac}^{\ell}\|
    \le
    \|\bar V^\ell V^{\ell+1}\|\,\|R_{\rm chain}^{\ell}\|
    +
    \epsilon_v^{\ell+1}\,\|\bar J_{\rm SAM}^{\ell}\|,
\end{equation}
since $\|I-P_{\ell+1}\|\le1$.

\begin{remark}[Keeping $\bar J_{\rm SAM}$ in the residual]\label{rem:keep-Jbar}
The error residual expressions (e.g., \eqref{eq:RJac-def} and \eqref{eq:RJac-bound}) retain $\bar J_{\rm SAM}^\ell$, a defined downstream product rather than a primitive error.
This is admissible because \eqref{eq:I6-entry-leakage} multiplies it by $E_v^{\ell+1}$: the contribution is $O\bigl(\epsilon_v^{\ell+1}\|\bar J_{\rm SAM}^\ell\|\bigr)$ and is therefore linear in that primitive error.
Lemma~\ref{lem:Ev} sends $E_v^{\ell+1}\to 0$ unconditionally, so the term vanishes without a separate smallness hypothesis on $\|\bar J_{\rm SAM}^\ell\|$ (bounded $\bar J_{\rm SAM}^\ell$ on the sample suffices).
\end{remark}

\medskip
\textbf{I6.6 (weight update).}
Apply \eqref{eq:I6-Jac} to $-g^L$. The right-hand side uses $g^{\ell+1}=J_{\rm SGD}^{\ell}g^L$ and $\bar a^L=-g^L$. The left-hand side is $J_{\rm SAM}^{\ell}\bar a^L$; with $J_{\rm SAM}^{\ell}=\bar V^\ell\bar J_{\rm SAM}^{\ell}$ and $\bar a^{\ell+1}=\bar J_{\rm SAM}^{\ell}\bar a^L$,
\[
    J_{\rm SAM}^{\ell}\bar a^L
    =
    \bar V^\ell\bar J_{\rm SAM}^{\ell}\bar a^L
    =
    \bar V^\ell\bar a^{\ell+1}.
\]
Hence
\begin{equation}\label{eq:I6-VJP}
    \bar V^\ell\bar a^{\ell+1}
    =
    -\bigl(H^\ell+\widetilde E_v^\ell\bigr)\,g^{\ell+1}
    - \mathcal{R}_{\rm Jac}^{\ell} g^L.
\end{equation}
The SAL and SGD updates on the same sample are
\[
    \Delta_{\rm SAL}W^\ell
    =
    \eta_W D_u^{\ell+1}\bar V^\ell\bar a^{\ell+1}(h^\ell)^T D_u^\ell-\gamma W^\ell,
    \qquad
    \Delta_{\rm sgd}W^\ell
    =
    -g^{\ell+1}(D_u^\ell h^\ell)^T .
\]
Since $D_u^T=D_u$, the right-hand factor of the SAL rule is $(h^\ell)^T D_u^\ell=(D_u^\ell h^\ell)^T$.
Insert \eqref{eq:I6-VJP}:
\begin{align*}
    D_u^{\ell+1}\bar V^\ell\bar a^{\ell+1}(D_u^\ell h^\ell)^T
    &=
    D_u^{\ell+1}
    \bigl[-\bigl(H^\ell+\widetilde E_v^\ell\bigr)g^{\ell+1}
          -\mathcal{R}_{\rm Jac}^{\ell}g^L\bigr]
    (D_u^\ell h^\ell)^T\\
    &=
    D_u^{\ell+1}\bigl(H^\ell+\widetilde E_v^\ell\bigr)
    \bigl(-g^{\ell+1}(D_u^\ell h^\ell)^T\bigr)
    -D_u^{\ell+1}(\mathcal{R}_{\rm Jac}^{\ell}g^L)(D_u^\ell h^\ell)^T\\
    &=
    D_u^{\ell+1}\bigl(H^\ell+\widetilde E_v^\ell\bigr)
    \Delta_{\rm sgd}W^\ell
    -D_u^{\ell+1}(\mathcal{R}_{\rm Jac}^{\ell}g^L)(D_u^\ell h^\ell)^T .
\end{align*}
For $\ell\le L-2$, exact autograd gives $D_u^{\ell+1}g^{\ell+1}=g^{\ell+1}$, hence
$D_u^{\ell+1}\Delta_{\rm sgd}W^\ell=\Delta_{\rm sgd}W^\ell$.
The principal term therefore sandwiches the right-hand gate:
\[
    D_u^{\ell+1}\bigl(H^\ell+\widetilde E_v^\ell\bigr)\Delta_{\rm sgd}W^\ell
    =
    D_u^{\ell+1}\bigl(H^\ell+\widetilde E_v^\ell\bigr)D_u^{\ell+1}\Delta_{\rm sgd}W^\ell
    =
    \bigl(H_D^\ell+\widetilde E_{v,D}^\ell\bigr)\Delta_{\rm sgd}W^\ell ,
\]
where
\begin{equation}\label{eq:HD-def}
    H_D^\ell:=D_u^{\ell+1}H^\ell D_u^{\ell+1},
    \qquad
    \widetilde E_{v,D}^\ell:=D_u^{\ell+1}\widetilde E_v^\ell D_u^{\ell+1},
\end{equation}
with $H^\ell$ and $\widetilde E_v^\ell$ from \eqref{eq:H-def}.
Multiply by $\eta_W$ and restore the decay:
\begin{equation}\label{eq:I6-main}
    \Delta_{\rm SAL}W^\ell
    =
    \eta_W\,\bigl(H_D^\ell+\widetilde E_{v,D}^\ell\bigr)
    \Delta_{\rm sgd}W^\ell
    - \gamma W^\ell + R^\ell,
\end{equation}
\[
    R^\ell
    =
    -\eta_W D_u^{\ell+1}
    (\mathcal{R}_{\rm Jac}^{\ell} g^L)
    (D_u^\ell h^\ell)^T.
\]
Combining \eqref{eq:RJac-bound} with submultiplicativity gives \eqref{eq:R-bound}.
Since $\|\mathcal{R}_{\rm Jac}^\ell\|$ is linear in the primitive errors and the local factors $\eta_W$, $D_u^{\ell+1}$, $g^L$, and $D_u^\ell h^\ell$ are bounded on the sample, $\|R^\ell\|$ is likewise linear in those errors.
This proves Lemma~\ref{lem:depth-chain} (Steps~I1--I5) and Theorem~\ref{thm:main} (Step~I6).

\section{Biological Evidence for Synaptic Plasticity Rules}
\label{Appendix-Plausibility-Rules} 

The learning rules corresponding to the SAM dynamics in the main text are not strictly Hebb. In fact, the classical Hebb rule is an example of homosynaptic plasticity, whereas the proposed algorithm requires the plasticity of neighboring synapses through heterosynaptic interactions. One of the first examples of heterosynaptic facilitation is in \cite{kandel1965}.
There is empirical evidence that reinforcement of a synapse based on its own input and nearby synaptic input can occur independently of large postsynaptic depolarization, often through localized biochemical signaling rather than direct electrical depolarization of the postsynaptic cell. In this case, synaptic plasticity is influenced by the activity of neighboring synapses without necessitating a strong postsynaptic action potential. A possible mechanism is Spike-Timing-Dependent Plasticity (STDP) with Local Modulation: in some forms of STDP, nearby synapses on the same dendrite can influence plasticity based on the timing of inputs. For example, the timing of presynaptic spikes at two nearby excitatory synapses can trigger local signaling cascades within the dendritic segment that reinforce or weaken synaptic strength without requiring a large postsynaptic potential. Another is calcium-dependent modulation: local increases in calcium concentration can drive synaptic reinforcement or weakening in a spatially restricted manner. When one synapse is active, it can cause a modest calcium influx that can extend to nearby synapses without significantly depolarizing the postsynaptic neuron. This local calcium increase can activate signaling pathways, such as CaMKII or other kinases, that modify the strength of nearby synapses proportionally to their own activity levels and those of neighboring inputs. A third possible mechanism is retrograde signaling, which involves retrograde messengers (e.g., endocannabinoids or nitric oxide) that are released by the postsynaptic neuron in response to local synaptic activity and can act on presynaptic terminals. These messengers can diffuse locally and affect only the nearby synapses that are also active, reinforcing or weakening them proportionally to their individual input activities. An interesting case study is by \cite{harvey2007} on synaptic tagging in hippocampal neurons. It demonstrated that synapses can interact biochemically on a local level. When a strong input induces LTP at one synapse, it can create a "tag" that allows nearby synapses to capture plasticity-related proteins and reinforce their strength based on their activity, even if they don't independently trigger large postsynaptic depolarization. These mechanisms allow synaptic connections to encode complex input patterns across local networks, effectively performing computations that go beyond simple, strict  Hebbian principles\footnote{There are a few different possibilities of how to implement them biologically. One special detail one needs to pay attention to is the time ordering of firing.}. 

In the main text, we have mainly focused on the simplest type of regularization. A more biologically plausible regularization is perhaps a Winner-Take-All regularization, and this could be a direction of future research.

\vspace{0.1in}
{\it Biological evidence for competition among incoming synapses on a single neuron (or dendrite).} Research does suggest a competitive process among synapses on the same dendritic branch, particularly when multiple inputs converge on a single location. This competition is thought to be governed by mechanisms of synaptic plasticity, where synapses that are frequently active and contribute to the cell’s firing are strengthened, while less active ones may weaken or even be pruned.
One well-documented example of this competition is seen in activity-dependent plasticity processes, such as Hebbian plasticity and synaptic tagging and capture. When neighboring synapses are active at the same time, they may compete for limited resources, like proteins that help stabilize and strengthen synaptic connections. This often results in the selective strengthening of certain synapses and the weakening of others, a process referred to as heterosynaptic plasticity. In this way, some synapses are favored to persist, while others may diminish over time if they fail to “compete” effectively. This competitive mechanism is believed to play a crucial role in refining neural circuits, helping to maintain efficient and effective synaptic networks by eliminating redundant or less useful connections, especially during development and learning.

\vspace{0.1in}
{\it Biological evidence for competition among synapses from a single neuron (or axon) onto different neurons.} There is evidence supporting the idea of presynaptic competition between synapses originating from the same neuron when they connect to different target neurons.
One of the most studied examples of presynaptic competition is in motor neurons and their connections to muscle fibers, especially during development. Motor neurons initially form multiple synapses on various muscle fibers, but over time, a pruning process occurs, often resulting in a single, stable connection with one muscle fiber. This phenomenon, known as {\it synaptic pruning}, is a process where less active synapses are eliminated while only the more effective and frequently active connections are maintained and strengthened.
This presynaptic competition is driven by neuronal activity and molecular signals. Research on long-term synaptic plasticity (LTP and LTD) suggests that competing synapses can affect each other’s likelihood of survival, even when they connect to different cells. Synchronized activity at certain synapses tends to stabilize those connections, while less synchronized or less active synapses are gradually weakened and may be removed.
Presynaptic competition is essential not only for refining neural networks but also for developing functional connections between neurons. It ensures that only the most efficient and relevant connections are maintained, optimizing neural circuitry for effective function.

\section{Gradient Learning Without a Two-Stream Scaffold}
\label{si:hsp-evolution}
 
At its core, SAM is a local heterosynaptic learning rule that makes a population
of neurons self-organize; the two interacting streams of the main text are one
particular setting in which we exercise this rule, not a defining commitment.
Companion work \cite{ziyin2025} applies the same rule in a much more general
setting---a generalized SAM---and asks how far the self-organization reaches when
the two-stream organization is removed entirely. A pool of neurons is randomly
initialized and recurrently connected, with no partition into separate streams;
every synapse follows the same local heterosynaptic rule; and an evolutionary
process grows and prunes the connections, selecting only on task performance. The
sole structure fixed in advance is where the error signal enters the circuit.
 
\begin{figure}[t!]
    \centering
    \includegraphics[width=\linewidth]{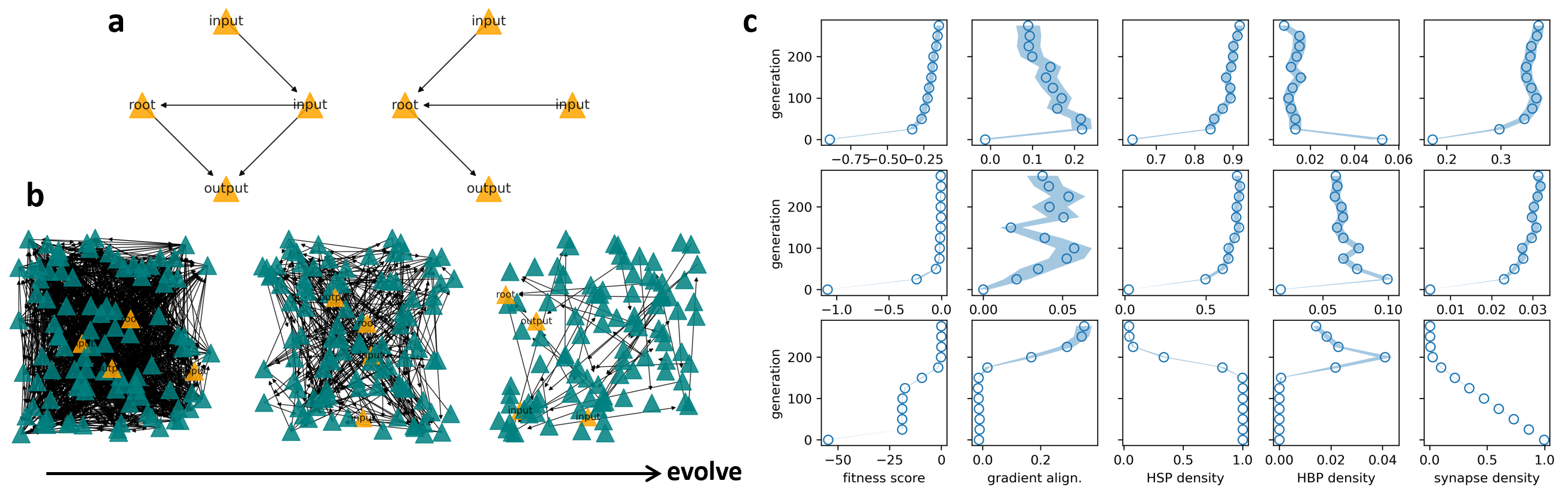}
    \caption{\textbf{Effective learning circuits self-assemble from a randomly
    initialized pool of neurons under the heterosynaptic rule, with no two-stream
    scaffold.} Reproduced from \cite{ziyin2025}.
    \textbf{(a)} Examples of evolved four-neuron circuits trained on a linear
    regression task; synapses are grown and pruned over generations.
    \textbf{(b)} Snapshots of an evolving $100$-neuron network.
    \textbf{(c)} Quantities versus generation (rows: four-neuron sparse init.,
    $100$-neuron sparse init., $100$-neuron dense init.), showing fitness,
    gradient alignment, heterosynaptic (HSP) density, Hebbian (HBP) density, and
    synapse density; shaded regions indicate variability across runs. As evolution
    proceeds, the evolved updates become increasingly aligned with the true
    gradient.}
    \label{fig:hsp-evolution}
\end{figure}

From this scaffold-free setting, effective learning circuits self-organize
(Fig.~\ref{fig:hsp-evolution}): driven by the heterosynaptic rule, the
connectivity reorganizes across generations into circuits that learn well, with
the evolved updates becoming increasingly aligned with the true gradient. Good
credit-assignment circuitry therefore self-assembles from a generic pool of
neurons, without any scaffold. We take this as evidence on two fronts: that the
heterosynaptic rule introduced in this work is powerful enough, on its own, to
make a population of neurons self-organize into an effective learner; and that the
SAL goal---learning structure emerging from local rules rather than from
design---is achievable across a broad range of settings, well beyond the
particular two-stream instantiation analyzed in the main text. In this sense the
result supports the general direction of SAL. We refer the reader to
\cite{ziyin2025} for the learning rule, the evolutionary algorithm, and the full
set of simulations.
 
\section{Experimental Setup}\label{app sec: experiment}

\begin{figure}[t!]
    \centering
    \IfFileExists{figures/acc_lr_heatmaps.pdf}{%
        \includegraphics[width=1\linewidth]{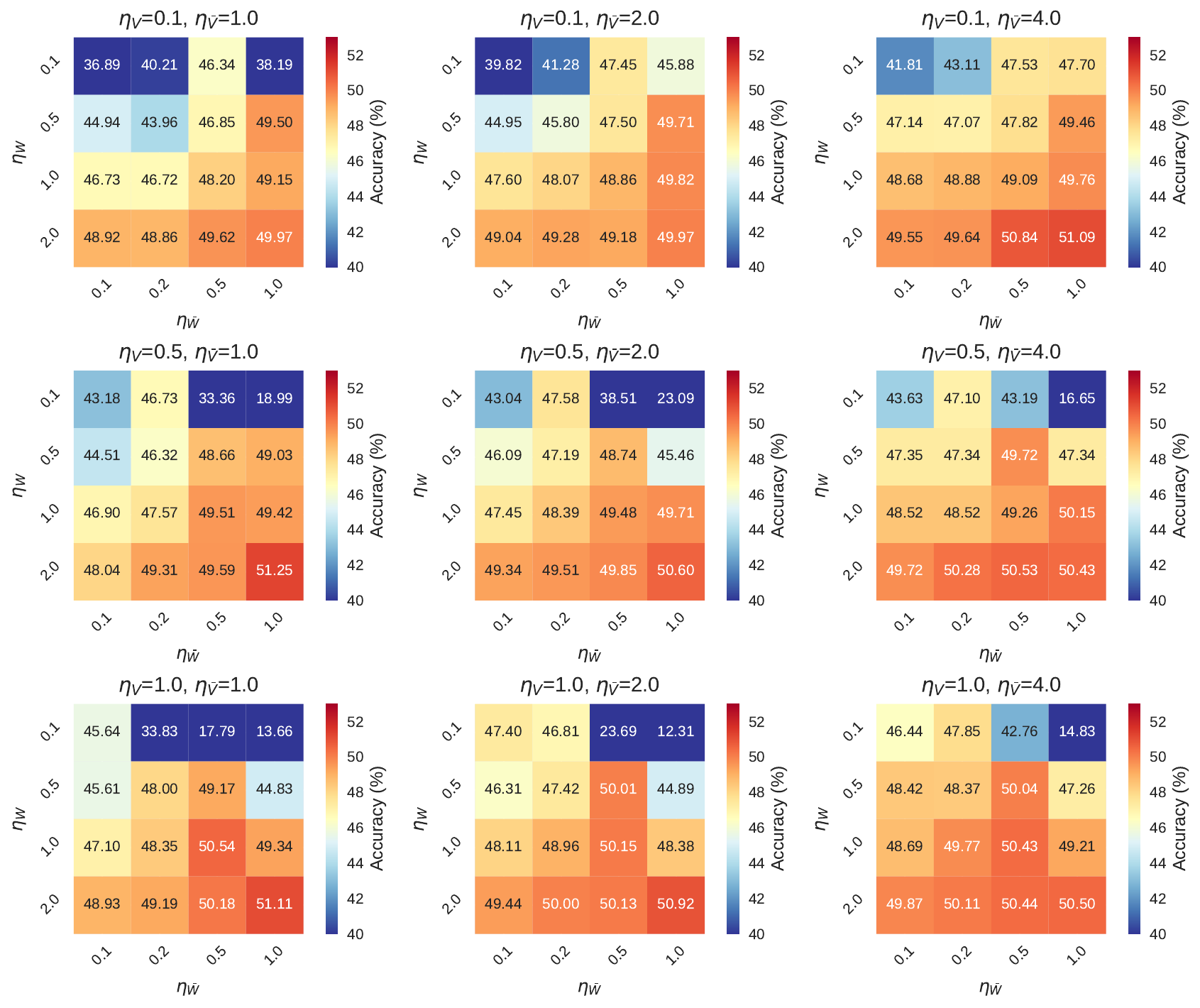}%
    }{\centering\fbox{\parbox{0.8\linewidth}{\centering\small [figures/acc\_lr\_heatmaps.pdf not found]}}}
    \caption{Performance of a four-layer MLP for different choices of learning rates. Interestingly, the best performance is achieved when $\eta_V$ is the smallest of the four learning rates.}
    \label{fig:learning rate grid search}
\end{figure}

All experiments are conducted with PyTorch on NVIDIA GPUs (mostly A100, occasionally L40S). The batch size is set to 256.

\paragraph{Datasets.} We use seven datasets to evaluate our method. \textit{CIFAR-10} \cite{krizhevsky2009} consists of 60,000 color images in 10 different classes, with each image having a resolution of 32x32 pixels. The dataset is divided into 50,000 training images and 10,000 testing images. The classes are mutually exclusive and include common objects such as airplanes, cars, cats, and dogs.

The \textit{MNIST} \cite{yann1998} contains 70,000 grayscale images of handwritten digits (0-9), each with a resolution of 28x28 pixels. The dataset is divided into 60,000 training images and 10,000 testing images. MNIST is known for its simplicity and has been a standard dataset for testing machine learning algorithms.

\textit{Fashion MNIST} \cite{xiao2017} contains 70,000 grayscale images of clothing items like T-shirts, trousers, shoes, and bags, with each image having a resolution of 28x28 pixels. The dataset is also split into 60,000 training and 10,000 testing images. Fashion MNIST is considered more challenging than MNIST due to the variability in the visual features of clothing items.

The \textit{SVHN} \cite{netzer2011} contains over 600,000 color images of digits (0-9), each with a resolution of 32x32 pixels. The dataset is divided into a training set of 73,257 images, a testing set of 26,032 images, and an additional 531,131 images for extra training. SVHN is challenging due to the varying digit sizes, orientations, and complex backgrounds.

The ChestMNIST \cite{yang2021,yang2023} comprises chest X-ray images for multi-label classification tasks. Derived from the NIH ChestX-ray14 dataset, it includes images annotated with 14 different pathological labels, such as pneumonia and emphysema. It is widely used in medical imaging studies to develop and evaluate machine learning models capable of diagnosing multiple conditions from X-ray images.

PathMNIST \cite{yang2021,yang2023} is a pathology image dataset designed for multi-class classification tasks. It originates from the NCT-CRC-HE-100K dataset and includes histological images of nine different types of tissue from colorectal cancer samples, such as colorectal adenocarcinoma epithelium and lymphocytes. This dataset is used primarily for training models to differentiate between various cancerous tissue types based on their histopathological features.

\textit{STL-10} \cite{coates2011} consists of 13,000 labeled images across 10 different classes and 100,000 unlabeled images. Each image has a resolution of 96x96 pixels, which is significantly larger than CIFAR-10 images, making the dataset more challenging. The dataset has a training set of 5,000 labeled images and a test set of 8,000 labeled images.

\paragraph{Baselines.}

We compare our method with four baselines, including SGD and three biologically-motivated algorithms.

\textit{Stochastic Gradient Descent (SGD)} is the standard optimization method that updates parameters iteratively using a subset of data to minimize the objective function.

\textit{Weight Mirroring} \cite{akrout2019a} learns feedback weights that approximate the transpose of the forward weights without weight transport: in a separate mirroring phase, noise is injected into the forward neurons and the feedback weights are updated with a local Hebbian rule, so they come to mirror the forward weights.

\textit{Feedback Alignment} \cite{lillicrap2016} offers an alternative to traditional backpropagation by using random, fixed weights instead of transposed forward-pass weights to propagate error signals.

\textit{Kolen-Pollack (KP) algorithm} \cite{kolen1994,akrout2019a} learns feedback weights without weight transport by applying the same local update (transposed) and the same weight decay to the forward and feedback weights during training, so their difference decays exponentially and the feedback weights converge to the transpose of the forward weights.

\section{Alignment Diagnostics}\label{sec:alignment-diagnostics}

This section reports CIFAR-10 SAL vs.\ backprop alignment diagnostics that test the theory in \S\ref{sec:theory}.
The runs follow Algorithm~\ref{alg:pseudocode-code2} with hyperparameters as below unless noted.

\paragraph{Two error budgets.}
The diagnostics separate two alignment quantities from \S\ref{sec:theory}.
\textbf{Interstream alignment} ($E_v^\ell\to 0$) is proved unconditionally when $\eta_V=\eta_{\bar V}$ (Lemma~\ref{lem:Ev}); any residual junction mismatch in simulations reflects finite training time, stochastic sampling, or numerical precision.
\textbf{Pathway alignment} ($E_p^\ell$ small) is conditional: convergence $E_p^\ell\to 0$ from the weight rules alone is not proved (Remark~\ref{rem:early-Ev}).
In stable runs we often observe $\|E_p^\ell\|$ and $\|E_p^\ell\|/\|\mathcal{F}^\ell\|$ decrease on most layers; when a run becomes unstable these quantities can instead grow.
When and why $E_p^\ell$ improves or worsens during training is left to future work.

\paragraph{Measured quantities.}
For each layer $\ell$ and sample $x$ we form the forward pathway operator $\mathcal{F}^\ell(x)=V^\ell D_u^\ell(x)(W^\ell)^T$, the feedback pathway operator $\mathcal{B}^\ell(x)=D_d^\ell(x)\bar W^\ell(\bar V^\ell)^T$, and the pathway error $E_p^\ell(x)=\mathcal{F}^\ell(x)-\mathcal{B}^\ell(x)$ (Definition~\ref{def:primitive-errors}).
Interstream alignment is proxied by junction mismatch $|V^{\ell+1}-\bar V^{\ell\mathrm{T}}|$ and cosine similarity $\cos(V^{\ell+1},\bar V^{\ell\mathrm{T}})$.
We also log SAL vs.\ backprop gradient cosine similarity on $W^\ell$, interstream gradient magnitudes $|\nabla V^\ell|$ and $|\nabla\bar V^\ell|$, per-layer $\|E_p^\ell\|_F$, and $\|E_p^\ell\|_F/\|\mathcal{F}^\ell\|_F$; since $E_p^\ell=\mathcal{F}^\ell-\mathcal{B}^\ell$, the ratio expresses operator mismatch relative to the forward pathway scale and is comparable across layers.
$V^0$ and $\bar W^0$ are unused by the model yet enter $E_p^0$, and a fuller treatment is left to future work.

\paragraph{Training hyperparameters.}
We train on CIFAR-10 with three or four hidden layers of width $200$ (readout index $L=4$ and $L=5$ respectively; $h^0$ is the input), ReLU forward nonlinearity, identity backward nonlinearity ($D_d=I$), no batch or layer normalization, batch size $256$, $50$ epochs, and alignment metrics recorded every five training steps.
Optimization uses SGD with base learning rate $\eta=5\times10^{-3}$ and momentum $0.9$.
Per-matrix SAL gradient scales are $\eta_W=1.0$, $\eta_{\bar W}=0.8$, and matched interstream rates $\eta_V=\eta_{\bar V}=0.8$.
Weight decay is $10^{-3}$ on $W^\ell$ and $\bar W^\ell$ and $0.01$ on $V^\ell$ and $\bar V^\ell$.
The 4-hidden-layer ($L=5$) divergent run differs only in depth.

\begin{figure*}[t!]
\centering
\IfFileExists{figures/alignment_3layer.pdf}{%
    \includegraphics[width=\textwidth]{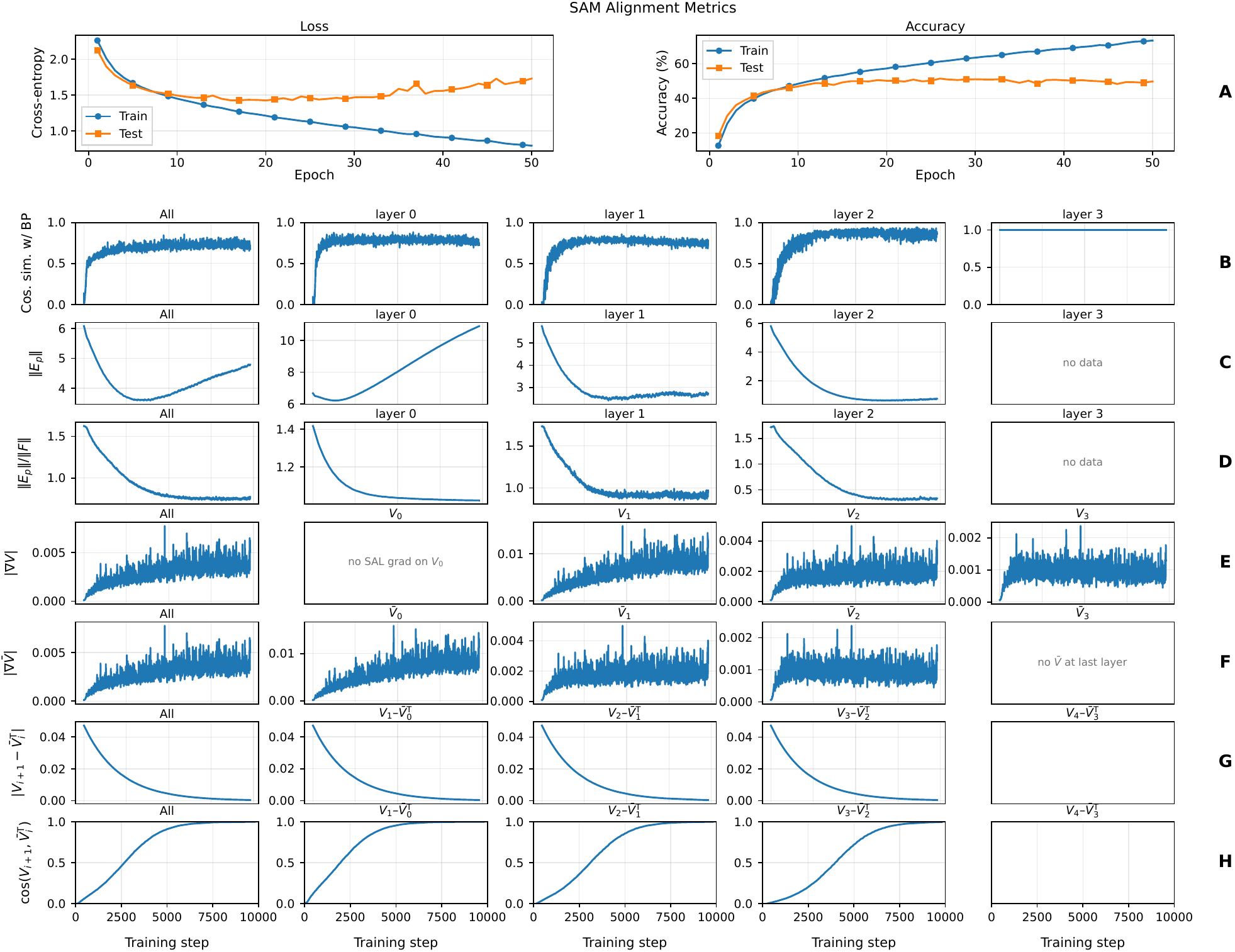}%
}{\centering\fbox{\parbox{0.8\linewidth}{\centering\small [figures/alignment\_3layer.pdf not found]}}}
\caption{\textbf{Alignment diagnostics on CIFAR-10 (3 hidden layers, $L=4$; stable run).}
SAL vs.\ backprop comparison during training (hyperparameters above).
\textbf{A:} train and test cross-entropy and accuracy over epochs.
\textbf{B:} cosine similarity between SAL and backprop gradients on $W^\ell$.
\textbf{C--D:} pathway error $\|E_p^\ell\|$ and $\|E_p^\ell\|/\|\mathcal{F}^\ell\|$ ($E_p^\ell=\mathcal{F}^\ell-\mathcal{B}^\ell$, normalized by $\|\mathcal{F}^\ell\|$; typically decreasing).
\textbf{E--F:} interstream gradient magnitudes $|\nabla V^\ell|$ and $|\nabla\bar V^\ell|$.
\textbf{G--H:} interstream mismatch $|V^{\ell+1}-\bar V^{\ell\mathrm{T}}|$ and cosine similarity.
Interstream self-assembly completes while plasticity remains active; pathway error falls on most layers; SAL tracks backprop on $W^\ell$. These are early diagnostics; a systematic analysis of $E_p$ is left to future work.}
\label{fig:si-alignment-3layer}
\end{figure*}

\paragraph{Three hidden layers ($L=4$; stable regime).}
With three hidden layers, training remains stable for the full $50$ epochs (Figure~\ref{fig:si-alignment-3layer}).
\textbf{Interstream alignment} (panels G--H): $|V^{\ell+1}-\bar V^{\ell\mathrm{T}}|$ falls and cosine similarity approaches one on all layers, even though $|\nabla V^\ell|$ and $|\nabla\bar V^\ell|$ in panels E--F stay at a stable positive level.
Transpose self-assembly ($E_v^\ell\to 0$) therefore completes while interstream \emph{plasticity} remains active---the central empirical distinction from requiring vanishing weight updates.
\textbf{Pathway alignment} (panels C--D): $\|E_p^\ell\|$ and $\|E_p^\ell\|/\|\mathcal{F}^\ell\|$ decrease on most layers; $E_p$ tracks pathway misalignment in practice once $E_v$ is small.
Together with tight interstream alignment, this is consistent with the conditional main theorem (Theorem~\ref{thm:main}), under which moderate pathway error is the remaining hypothesis for SAL gradients on $W^\ell$ to track a preconditioned backprop direction (panel~B).

\begin{figure*}[t!]
\centering 
\IfFileExists{figures/alignment_4layer.pdf}{%
    \includegraphics[width=\textwidth]{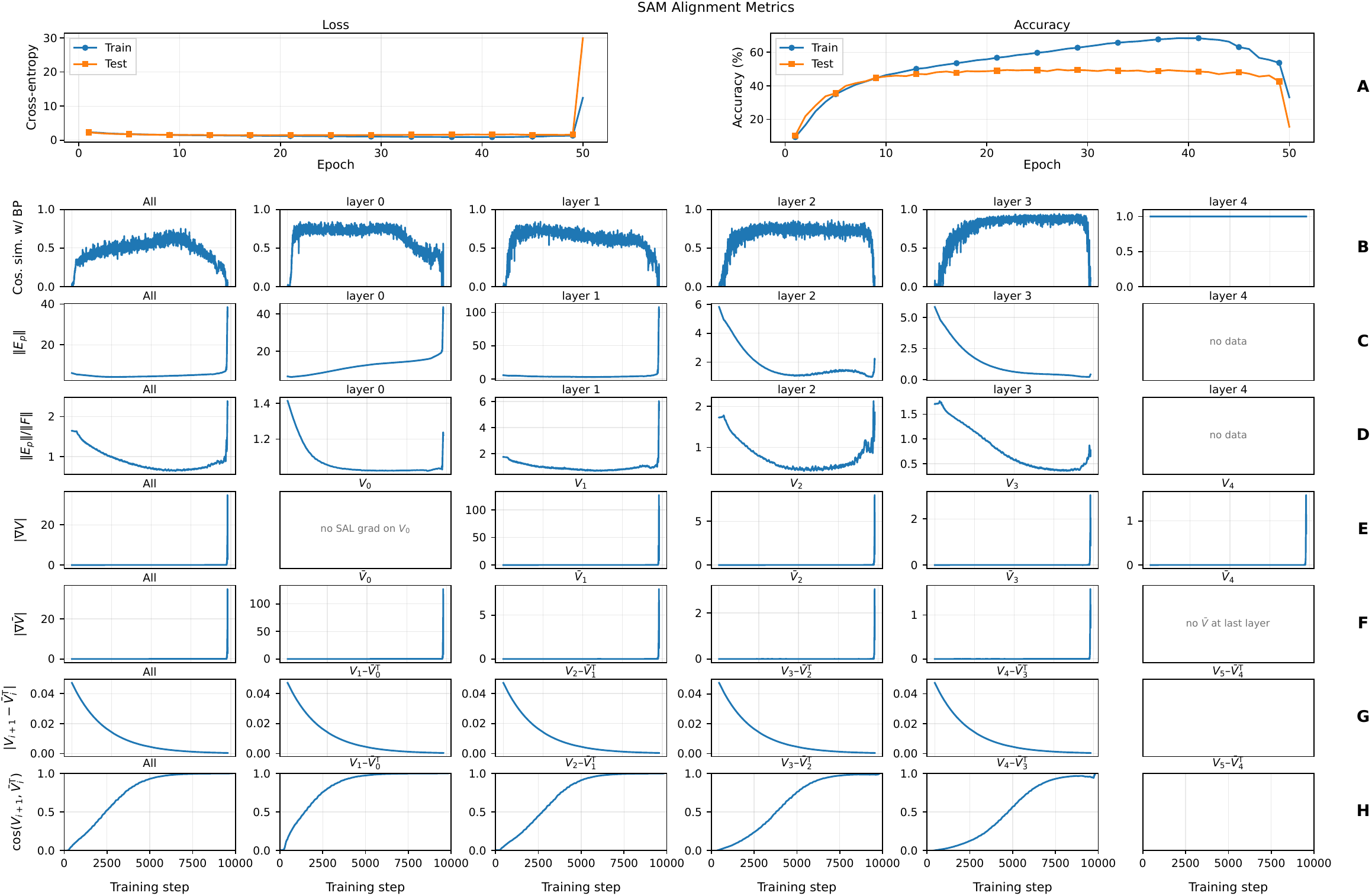}%
}{\centering\fbox{\parbox{0.8\linewidth}{\centering\small [figures/alignment\_4layer.pdf not found]}}}
\caption{\textbf{Alignment diagnostics on CIFAR-10 (4 hidden layers, $L=5$; deliberately chosen divergent run).}
Same panel layout and default hyperparameters as Figure~\ref{fig:si-alignment-3layer}, with four hidden layers; divergence is not inevitable at this depth and depends on learning-rate tuning.
$\|E_p^\ell\|$ can decrease on some layers early, while rising around the time instability appears; interstream alignment in panels G--H can remain good on all layers when failure happens.}
\label{fig:si-alignment-4layer}
\end{figure*}

\paragraph{Four hidden layers ($L=5$; illustrative divergent run).}
With four hidden layers and the same default hyperparameters as above, training need not diverge: stability depends on the relative learning rates of $W$, $\bar W$, $V$, and $\bar V$ and on the overall learning-rate scale, and can often be restored by tuning them.
We deliberately report one divergent run (Figure~\ref{fig:si-alignment-4layer}) to illustrate that $E_p$ and $E_v$ can behave differently when training becomes unstable.
Early in training this run can resemble the 3-hidden-layer ($L=4$) case: loss decreases, V-alignment improves, and $\|E_p^\ell\|$ on some layers often decreases.
When the run diverges, $\|E_p^\ell\|$ can rise on some layers, even if it had been decreasing earlier.
Panels G--H show interstream alignment remains good on all layers when divergence begins, so the failure is not accompanied by a collapse of $E_v$.
This is compatible with the theorem's separation of the two error budgets.
A theory of when and why $E_p$ grows or decreases remains to be developed.
   
\section{Pseudo Code}\label{app sec: algorithm}

We give two algorithmic presentations of SAM, corresponding to two code releases on GitHub.
Algorithm~\ref{alg:pseudocode-code1} is the \textbf{2024/2025} training loop used for the benchmark experiments in the main text (Table~1, Figure~3 ablations, width study, and related runs).
Algorithm~\ref{alg:pseudocode-code2} is the \textbf{2026} implementation: it matches the local plasticity rules in SI Theory (\S\ref{sec:theory}, \eqref{eq:dV}--\eqref{eq:dbarV} and \eqref{eq:I6-main}), and the alignment diagnostics in \S\ref{sec:alignment-diagnostics} use this formulation.
The two presentations differ in layer indexing, explicit ReLU gates ($D_u$, $D_d$), post-activation ports, and per-matrix learning rates; benchmark runs use $D_d=I$ on the downstream pathway (take $D=I$ in the backward pass of Algorithm~\ref{alg:pseudocode-code1}).
Algorithm~\ref{alg:pseudocode-code1} is the original benchmark implementation, which omits the postsynaptic nonlinearity on $V$ and $\bar V$ updates.
Algorithm~\ref{alg:pseudocode-code2} is the more symmetric gated form used for the alignment-based theory and diagnostics.
We did not observe qualitatively different behavior between the two variants. In Algorithm~\ref{alg:pseudocode-code2}, the paired $V^\ell$ and $(\bar V^{\ell-1})^T$ updates share the same outer-product drive, so Lemma~\ref{lem:Ev} applies when $\eta_V=\eta_{\bar V}$; in Algorithm~\ref{alg:pseudocode-code1}, the two drives differ only by a term involving the negative part of the pre-activation, so the same mechanism applies approximately.

\begin{algorithm}[t]
\caption{SAM training loop (\textbf{2024/2025} benchmark implementation; layer index starts at $1$)}\label{alg:pseudocode-code1}
\begin{algorithmic}[1]
\State \textbf{Initialize:} For each layer $\ell = 1$ to $L$, initialize weight matrices $W^\ell$, $V^\ell$, $\bar{W}^\ell$, and $\bar{V}^\ell$
\For{each training example}
    \State \textbf{Forward Pass (Upstream Pathway):}
    \State $h^1 \gets$ input data $x$
    \For{$\ell = 1$ to $L-1$}
        \State $h^{\ell+1} \gets D(W^\ell h^\ell)$
    \EndFor
    \State \textbf{Compute Loss:}
    \State Compute loss $F$ based on output $h^{L}$ and target $y$
    \State \textbf{Initialize Error Signal:}
    \State $\bar{h}^{L} \gets \epsilon(h^{L}, y)$ \hfill (e.g., $\bar{h}^{L} = -\partial F/\partial h^{L}$)
    \State \textbf{Backward Pass (Downstream Pathway):}
    \For{$\ell = L-1$ down to $1$}
        \State $\bar{h}^\ell \gets D(\bar{W}^\ell \bar{h}^{\ell+1})$
    \EndFor
    \State \textbf{Update Synaptic Weights:}
    \For{$\ell = 1$ to $L$}
        \State $\Delta W^\ell \gets \bar{V}^\ell \bar{h}^{\ell+1} (h^\ell)^\top - \gamma W^\ell$
        \State $\Delta V^\ell \gets \bar{W}^\ell \bar{h}^{\ell+1} (h^\ell)^\top - \gamma V^\ell$
        \State $\Delta \bar{W}^\ell \gets \left( \bar{h}^{\ell+1} (h^\ell)^\top (V^\ell)^\top - \gamma (\bar{W}^\ell)^\top \right)^\top$
        \State $\Delta \bar{V}^\ell \gets \left( \bar{h}^{\ell+1} (h^\ell)^\top (W^\ell)^\top - \gamma (\bar{V}^\ell)^\top \right)^\top$
        \State $W^\ell \gets W^\ell + \eta \Delta W^\ell$
        \State $V^\ell \gets V^\ell + \eta \Delta V^\ell$
        \State $\bar{W}^\ell \gets \bar{W}^\ell + \eta \Delta \bar{W}^\ell$
        \State $\bar{V}^\ell \gets \bar{V}^\ell + \eta \Delta \bar{V}^\ell$
    \EndFor
    \State At $\ell=1$, omit $\Delta V^1$ and $\Delta\bar{W}^1$; at $\ell=L$, omit $\Delta\bar{V}^L$
\EndFor
\end{algorithmic}
\end{algorithm}

\begin{algorithm}[t]
\caption{SAM training loop (\textbf{2026} implementation; SI Theory; layer index starts at $0$; the network does not grow or lose a layer relative to Algorithm~\ref{alg:pseudocode-code1})}\label{alg:pseudocode-code2}
\begin{algorithmic}[1]
\State \textbf{Initialize:} For each motif $\ell=0,\ldots,L-1$, initialize $W^\ell$, $V^\ell$, $\bar W^\ell$, $\bar V^\ell$
\For{each training example $(x,y)$}
    \State \textbf{Forward pass (upstream):}
    \State $h^0 \gets x$
    \For{$\ell = 0$ to $L-1$}
        \State $D_u^\ell \gets I$ if $\ell=0$, else $\diag(\mathbf{1}_{h^\ell>0})$ \hfill (input gate is identity)
        \State $a^\ell \gets D_u^\ell h^\ell$ \hfill (upstream activity port)
        \State $h^{\ell+1} \gets W^\ell a^\ell$
    \EndFor
    \State $D_u^L \gets I$;\quad $a^L \gets h^L$ \hfill (linear readout port)
    \State \textbf{Loss and readout cotangent:}
    \State Compute loss $F$ from output $h^L$ and label $y$
    \State $g^L \gets \partial F/\partial h^L$ \hfill (e.g.\ MSE $F=\|y-h^L\|^2$: $g^L = 2(h^L-y)$)
    \State \textbf{Backward pass (downstream activity port):}
    \State $\bar a^L \gets -g^L$
    \For{$\ell = L-1$ down to $0$}
        \State $\bar h^\ell \gets \bar W^\ell \bar a^{\ell+1}$
        \State $D_d^\ell \gets \diag(\mathbf{1}_{\bar h^\ell>0})$ \hfill (or $D_d^\ell \gets I$ in default runs)
        \State $\bar a^\ell \gets D_d^\ell \bar h^\ell$
    \EndFor
    \State \textbf{Heterosynaptic updates} (local decay $\gamma>0$; rates $\eta_W,\eta_{\bar W},\eta_V,\eta_{\bar V}$)
    \For{$\ell = 0$ to $L-1$}
        \State $\Delta W^\ell \gets \eta_W\, D_u^{\ell+1}\bar V^\ell \bar a^{\ell+1}\,(a^\ell)^T - \gamma W^\ell$
        \State $\Delta \bar W^\ell \gets \eta_{\bar W}\, D_d^\ell V^\ell a^\ell\,(\bar a^{\ell+1})^T - \gamma \bar W^\ell$
        \State $\Delta V^\ell \gets \eta_V\, \bar a^\ell (a^\ell)^T - \gamma V^\ell$
        \State $\Delta \bar V^\ell \gets \eta_{\bar V}\, a^{\ell+1}(\bar a^{\ell+1})^T - \gamma \bar V^\ell$
        \State $W^\ell \gets W^\ell + \Delta W^\ell$;\;
               $\bar W^\ell \gets \bar W^\ell + \Delta \bar W^\ell$
        \State $V^\ell \gets V^\ell + \Delta V^\ell$;\;
               $\bar V^\ell \gets \bar V^\ell + \Delta \bar V^\ell$
    \EndFor
    \State At $\ell=0$, omit $\Delta V^0$ and $\Delta\bar W^0$; at $\ell=L-1$, omit $\Delta\bar V^{L-1}$
\EndFor
\end{algorithmic}
\end{algorithm}

\clearpage 

\bibliography{pnas_ref_corrected}

@article{abel2024,
  title={Biologically inspired learning model for instructed vision},
  author={Abel, Roy and Ullman, Shimon},
  journal={Advances in Neural Information Processing Systems},
  volume={37},
  pages={45315--45358},
  year={2024},
  doi={10.52202/079017-1440}
}

@article{abraham1996,
  author = {Abraham, W. C. and Bear, M. F.},
  title = {Metaplasticity: the plasticity of synaptic plasticity},
  journal = {Trends in Neurosciences},
  volume = {19},
  number = {4},
  pages = {126--130},
  year = {1996}
}

@article{abraham2008,
  author = {Abraham, W. C.},
  title = {Metaplasticity: tuning synapses and networks for plasticity},
  journal = {Nature Reviews Neuroscience},
  volume = {9},
  number = {5},
  pages = {387},
  year = {2008}
}

@inproceedings{akrout2019a,
  author    = {Akrout, M. and Wilson, C. and Humphreys, P. C. and Lillicrap, T. P. and Tweed, D. B.},
  title     = {Deep learning without weight transport},
  booktitle = {Advances in Neural Information Processing Systems},
  editor    = {Wallach, H. and Larochelle, H. and Beygelzimer, A. and d'Alch\'e-Buc, F. and Fox, E. and Garnett, R.},
  publisher = {Curran Associates, Inc.},
  address   = {Red Hook, NY},
  volume    = {32},
  year      = {2019}
}

@article{anderson2009,
  author = {Anderson, J. C. and Martin, K. A.},
  title = {The synaptic connections between cortical areas {V1} and {V2} in macaque monkey},
  journal = {Journal of Neuroscience},
  volume = {29},
  number = {36},
  pages = {11283--11293},
  year = {2009}
}

@article{briggs2020,
  author = {Briggs, F.},
  title = {Role of feedback connections in central visual processing},
  journal = {Annual Review of Vision Science},
  volume = {6},
  number = {1},
  pages = {313--334},
  year = {2020}
}

@article{cichy2016,
  author = {Cichy, R. M. and Khosla, A. and Pantazis, D. and Torralba, A. and Oliva, A.},
  title = {Comparison of deep neural networks to spatio-temporal cortical dynamics of human visual object recognition reveals hierarchical correspondence},
  journal = {Sci. Rep.},
  volume = {6},
  pages = {27755},
  year = {2016}
}

@inproceedings{coates2011,
  author    = {Coates, A. and Ng, A. and Lee, H.},
  title     = {An analysis of single-layer networks in unsupervised feature learning},
  booktitle = {Proceedings of the Fourteenth International Conference on Artificial Intelligence and Statistics},
  editor    = {Gordon, Geoffrey and Dunson, David and Dud\'ik, Miroslav},
  series    = {Proceedings of Machine Learning Research},
  volume    = {15},
  publisher = {PMLR},
  pages     = {215--223},
  year      = {2011}
}

@article{guclu2015,
  author = {Güçlü, U. and van Gerven, M. A.},
  title = {Deep neural networks reveal a gradient in the complexity of neural representations across the ventral stream},
  journal = {Journal of Neuroscience},
  volume = {35},
  number = {27},
  pages = {10005--10014},
  year = {2015}
}

@article{harvey2007,
  author = {Harvey, C. D. and Svoboda, K.},
  title = {Locally dynamic synaptic learning rules in pyramidal neuron dendrites},
  journal = {Nature},
  volume = {450},
  number = {7173},
  pages = {1195--1200},
  year = {2007}
}

@article{huh2021,
  author  = {Huh, Minyoung and Mobahi, Hossein and Zhang, Richard and Cheung, Brian and Agrawal, Pulkit and Isola, Phillip},
  title   = {The low-rank simplicity bias in deep networks},
  journal = {Trans. Mach. Learn. Res.},
  issn    = {2835-8856},
  year    = {2023},
  url     = {https://openreview.net/forum?id=bCiNWDmlY2}
}

@article{kandel1965,
  author = {Kandel, E. R. and Tauc, L.},
  title = {Heterosynaptic facilitation in neurones of the abdominal ganglion of aplysia depilans},
  journal = {Journal of Physiology},
  volume = {181},
  number = {1},
  pages = {1--27},
  year = {1965},
  doi = {https://doi.org/10.1113/jphysiol.1965.sp007742},
  url = {https://physoc.onlinelibrary.wiley.com/doi/abs/10.1113/jphysiol.1965.sp007742}
}

@inproceedings{kolen1994,
  author    = {Kolen, J. and Pollack, J.},
  title     = {Back-propagation without weight transport},
  booktitle = {Proceedings of 1994 IEEE International Conference on Neural Networks (ICNN'94)},
  publisher = {IEEE},
  address   = {Piscataway, NJ},
  volume    = {3},
  pages     = {1375--1380},
  year      = {1994},
  doi       = {10.1109/icnn.1994.374486}
}

@article{kriegeskorte2015,
  author = {Kriegeskorte, N.},
  title = {Deep neural networks: A new framework for modeling biological vision and brain information processing},
  journal = {Annual Review of Vision Science},
  volume = {1},
  pages = {417--446},
  year = {2015}
}

@techreport{krizhevsky2009,
  author      = {Krizhevsky, A.},
  title       = {Learning multiple layers of features from tiny images},
  institution = {University of Toronto},
  address     = {Toronto, ON},
  year        = {2009},
  url         = {https://www.cs.toronto.edu/~kriz/learning-features-2009-TR.pdf}
}

@inproceedings{liao2015,
  author    = {Liao, Q. and Leibo, J. and Poggio, T.},
  title     = {How important is weight symmetry in backpropagation?},
  booktitle = {Proceedings of the Thirtieth AAAI Conference on Artificial Intelligence (AAAI-16)},
  editor    = {Schuurmans, Dale and Wellman, Michael P.},
  publisher = {AAAI Press},
  address   = {Palo Alto, CA},
  year      = {2016},
  note      = {arXiv:1510.05067},
  doi       = {10.1609/aaai.v30i1.10279}
}

@article{lillicrap2016,
  author = {Lillicrap, T. P. and Cownden, D. and Tweed, D. B. and Akerman, C. J.},
  title = {Random synaptic feedback weights support error backpropagation for deep learning},
  journal = {Nature Communications},
  volume = {7},
  number = {1},
  pages = {13276},
  year = {2016}
}

@article{lillicrap2020,
  author = {Lillicrap, T. P. and Santoro, A. and Marris, L. and Akerman, C. J. and Hinton, G.},
  title = {Backpropagation and the brain},
  journal = {Nature Reviews Neuroscience},
  volume = {21},
  number = {6},
  pages = {335--346},
  year = {2020}
}

@article{max2024,
  author = {Max, K. and Kriener, L. and García, G. P. and Nowotny, T. and Jaras, I. and Senn, W. and Petrovici, M. A.},
  title = {Learning efficient backprojections across cortical hierarchies in real time},
  journal = {Nature Machine Intelligence},
  volume = {6},
  number = {6},
  pages = {619--630},
  year = {2024},
  url = {http://dblp.uni-trier.de/db/journals/natmi/natmi6.html#MaxKGNJSP24}
}

@inproceedings{netzer2011,
  author    = {Netzer, Y. and Wang, T. and Coates, A. and Bissacco, A. and Wu, B. and Ng, A. Y.},
  title     = {Reading digits in natural images with unsupervised feature learning},
  booktitle = {NIPS Workshop on Deep Learning and Unsupervised Feature Learning},
  address   = {Granada, Spain},
  year      = {2011}
}

@inproceedings{nokland2016,
  author    = {N{\o}kland, A.},
  title     = {Direct feedback alignment provides learning in deep neural networks},
  booktitle = {Advances in Neural Information Processing Systems},
  editor    = {Lee, D. D. and Sugiyama, M. and von Luxburg, U. and Guyon, I. and Garnett, R.},
  publisher = {Curran Associates, Inc.},
  address   = {Red Hook, NY},
  volume    = {29},
  year      = {2016}
}

@inproceedings{nokland2019,
  author    = {N{\o}kland, A. and Eidnes, L. H.},
  title     = {Training neural networks with local error signals},
  booktitle = {Proceedings of the 36th International Conference on Machine Learning},
  editor    = {Chaudhuri, Kamalika and Salakhutdinov, Ruslan},
  series    = {Proceedings of Machine Learning Research},
  volume    = {97},
  publisher = {PMLR},
  pages     = {4839--4850},
  year      = {2019}
}

@inproceedings{roy2007,
  author    = {Roy, O. and Vetterli, M.},
  title     = {The effective rank: A measure of effective dimensionality},
  booktitle = {2007 15th European Signal Processing Conference (EUSIPCO)},
  publisher = {IEEE},
  address   = {Piscataway, NJ},
  pages     = {606--610},
  year      = {2007}
}

@article{sun2025,
  author = {Sun, W. and Winnubst, J. and Natrajan, M. and Lai, C. and Kajikawa, K. and Bast, A. and Michaelos, M. and Gattoni, R. and Stringer, C. and Flickinger, D. and Fitzgerald, J. E. and Spruston, N.},
  title = {Learning produces an orthogonalized state machine in the hippocampus},
  journal = {Nature},
  volume = {640},
  number = {8057},
  pages = {165--175},
  year = {2025}
}

@article{whittington2019,
  author = {Whittington, J. C. and Bogacz, R.},
  title = {Theories of error back-propagation in the brain},
  journal = {Trends in Cognitive Sciences},
  volume = {23},
  number = {3},
  pages = {235--250},
  year = {2019}
}

@article{xiao2017,
  author = {Xiao, H. and Rasul, K. and Vollgraf, R.},
  title = {{Fashion-MNIST}: a novel image dataset for benchmarking machine learning algorithms},
  journal = {arXiv preprint arXiv:1708.07747},
  year = {2017},
  note      = {Accessed 31/08/2026}
}

@inproceedings{xiao2018,
  author    = {Xiao, W. and Chen, H. and Liao, Q. and Poggio, T.},
  title     = {Biologically-plausible learning algorithms can scale to large datasets},
  booktitle = {International Conference on Learning Representations (ICLR)},
  year      = {2019},
  note      = {arXiv:1811.03567}
}

@article{yamins2016,
  author = {Yamins, D. L. and DiCarlo, J. J.},
  title = {Using goal-driven deep learning models to understand sensory cortex},
  journal = {Nature Neuroscience},
  volume = {19},
  number = {3},
  pages = {356--365},
  year = {2016}
}

@inproceedings{yang2021,
  author    = {Yang, J. and Shi, R. and Ni, B.},
  title     = {{MedMNIST} classification decathlon: A lightweight {AutoML} benchmark for medical image analysis},
  booktitle = {2021 IEEE 18th International Symposium on Biomedical Imaging (ISBI)},
  publisher = {IEEE},
  address   = {Piscataway, NJ},
  pages     = {191--195},
  year      = {2021},
  doi       = {10.1109/isbi48211.2021.9434062}
}

@article{yang2023,
  author = {Yang, J. and Shi, R. and Wei, D. and Liu, Z. and Zhao, L. and Ke, B. and Pfister, H. and Ni, B.},
  title = {{MedMNIST} v2---{A} large-scale lightweight benchmark for {2D} and {3D} biomedical image classification},
  journal = {Scientific Data},
  volume = {10},
  number = {1},
  pages = {41},
  year = {2023}
}

@misc{yann1998,
  author       = {LeCun, Y. and Cortes, C. and Burges, C. J. C.},
  title        = {The {MNIST} database of handwritten digits},
  howpublished = {Online database},
  year         = {1998},
  url          = {http://yann.lecun.com/exdb/mnist/}
}

@inproceedings{ziyin2024,
  author    = {Ziyin, Liu and Chuang, Isaac and Galanti, Tomer and Poggio, Tomaso},
  title     = {Formation of representations in neural networks},
  booktitle = {The Thirteenth International Conference on Learning Representations},
  year      = {2025}
}

@article{ziyin2025,
  author  = {Ziyin, Liu and Chuang, Isaac and Poggio, Tomaso},
  title   = {Heterosynaptic Circuits Are Universal Gradient Machines},
  journal = {arXiv preprint arXiv:2505.02248},
  year    = {2025},
  note      = {Accessed 31/08/2026}
}

@article{raugel2025,
  author = {Raugel, Joséphine and Szafraniec, Marc and Vo, Huy V. and Couprie, Camille and Labatut, Patrick and Bojanowski, Piotr and Wyart, Valentin and King, Jean-Rémi},
  title = {Disentangling the Factors of Convergence between Brains and Computer Vision Models},
  journal = {arXiv preprint arXiv:2508.18226},
  year = {2025},
  note      = {Accessed 31/08/2026}
}

@article{rosenbaum2023,
  author = {Shervani-Tabar, Navid and Rosenbaum, Robert},
  title = {Meta-learning biologically plausible plasticity rules with random feedback pathways},
  journal = {Nature Communications},
  volume = {14},
  number = {1},
  pages = {1805},
  year = {2023}
}

@article{friston2023,
  author = {Isomura, Takuya and Kotani, Kiyoshi and Jimbo, Yasuhiko and Friston, Karl J.},
  title = {Experimental validation of the free-energy principle with in vitro neural networks},
  journal = {Nature Communications},
  volume = {14},
  number = {1},
  pages = {4547},
  year = {2023}
}

@misc{sejnowski2025,
  author       = {Cameron, Mia and Chen, Yusi and Sejnowski, Terrence J.},
  title        = {A Biologically-Plausible Alternative to Backpropagation Using Pseudoinverse Feedback},
  howpublished = {OpenReview preprint, \url{https://openreview.net/forum?id=B5Dj4EhZPP}},
  year         = {2025},
  note      = {Accessed 31/08/2026}
}

@article{fu2011experience,
  author = {Fu, M. and Zuo, Y.},
  title = {Experience-dependent structural plasticity in the cortex},
  journal = {Trends in Neurosciences},
  volume = {34},
  number = {4},
  pages = {177--187},
  year = {2011}
}

@article{larkum2013cellular,
  author = {Larkum, M. E.},
  title = {A cellular mechanism for cortical associations: an organizing principle for the cerebral cortex},
  journal = {Trends in Neurosciences},
  volume = {36},
  number = {3},
  pages = {141--151},
  year = {2013}
}

@article{tattikota2017distinct,
  title = {Pyramidal neurons in different cortical layers exhibit distinct dynamics and plasticity of apical dendritic spines},
  author = {Tjia, M. and Yu, X. and Jammu, L. S. and Lu, J. and Zuo, Y.},
  journal = {Frontiers in Neural Circuits},
  volume = {11},
  pages = {43},
  year = {2017},
  doi = {10.3389/fncir.2017.00043}
}

@article{yang2009stably,
  title = {Stably maintained dendritic spines are associated with lifelong memories},
  author = {Yang, Guang and Pan, Feng and Gan, Wen-Biao},
  journal = {Nature},
  volume = {462},
  number = {7275},
  pages = {920--924},
  year = {2009},
  doi = {10.1038/nature08577}
}

@article{xu2009rapid,
  title = {Rapid formation and selective stabilization of synapses for enduring motor memories},
  author = {Xu, Tonghui and Yu, Xinzhu and Perlik, Andrew J. and Tobin, Willie F. and Zweig, Jonathan A. and Tennant, Kelly and Jones, Theresa and Zuo, Yi},
  journal = {Nature},
  volume = {462},
  number = {7275},
  pages = {915--919},
  year = {2009},
  doi = {10.1038/nature08389}
}

@article{holtmaat2005transient,
  title = {Transient and persistent dendritic spines in the neocortex in vivo},
  author = {Holtmaat, Anthony J. G. D. and Trachtenberg, Joshua T. and Wilbrecht, Linda and Shepherd, Gordon M. G. and Zhang, Xiaoqun and Knott, Graham W. and Svoboda, Karel},
  journal = {Neuron},
  volume = {45},
  number = {2},
  pages = {279--291},
  year = {2005},
  doi = {10.1016/j.neuron.2005.01.003}
}

@article{makino2015learning,
  title = {Learning enhances the relative impact of top-down processing in the visual cortex},
  author = {Makino, Hiroshi and Komiyama, Takaki},
  journal = {Nature Neuroscience},
  volume = {18},
  number = {8},
  pages = {1116--1122},
  year = {2015},
  doi = {10.1038/nn.4061}
}

@article{abs2018learning,
  title = {Learning-related plasticity in dendrite-targeting layer 1 interneurons},
  author = {Abs, E. and Poorthuis, R. B. and Apelblat, D. and Muhammad, K. and Pardi, M. B. and Enke, L. and Kushinsky, D. and Pu, D. L. and Eizinger, M. F. and Conzelmann, K. K. and Spiegel, I. and Letzkus, J. J.},
  journal = {Neuron},
  volume = {100},
  number = {3},
  pages = {684--699.e6},
  year = {2018},
  doi = {10.1016/j.neuron.2018.09.001}
}

@article{doron2020perirhinal,
  title = {Perirhinal input to neocortical layer 1 controls learning},
  author = {Doron, Guy and Shin, Jiyun N. and Takahashi, Naoya and Dr{\"u}ke, Moritz and Bocklisch, Christina and Skenderi, Salina and de Mont, Lisa and Toumazou, Maria and Ledderose, Joshua and Brecht, Michael and Naud, Richard and Larkum, Matthew E.},
  journal = {Science},
  volume = {370},
  number = {6523},
  pages = {eaaz3136},
  year = {2020},
  doi = {10.1126/science.aaz3136}
}

@article{sacramento2018dendritic,
  title={Dendritic cortical microcircuits approximate the backpropagation algorithm},
  author={Sacramento, Jo{\~a}o and Ponte Costa, Rui and Bengio, Yoshua and Senn, Walter},
  journal={Advances in Neural Information Processing Systems},
  volume={31},
  pages={8721--8732},
  year={2018}
}

@article{payeur2021burst,
  title={Burst-dependent synaptic plasticity can coordinate learning in hierarchical circuits},
  author={Payeur, Alexandre and Guerguiev, Jordan and Zenke, Friedemann and Richards, Blake A and Naud, Richard},
  journal={Nature Neuroscience},
  volume={24},
  number={7},
  pages={1010--1019},
  year={2021},
  publisher={Nature Publishing Group US New York}
}

@article{greedy2022single,
  title={Single-phase deep learning in cortico-cortical networks},
  author={Greedy, Will and Zhu, Heng Wei and Pemberton, Joseph and Mellor, Jack and Ponte Costa, Rui},
  journal={Advances in Neural Information Processing Systems},
  volume={35},
  pages={24213--24225},
  year={2022},
  doi={10.52202/068431-1758}
}

@article{driscoll2024,
  title   = {Flexible multitask computation in recurrent networks utilizes shared dynamical motifs},
  author  = {Driscoll, Laura N. and Shenoy, Krishna and Sussillo, David},
  journal = {Nature Neuroscience},
  volume  = {27},
  number  = {7},
  pages   = {1349--1363},
  year    = {2024},
  doi     = {10.1038/s41593-024-01668-6}
}

@article{kim2021,
  title   = {Strong inhibitory signaling underlies stable temporal dynamics and working memory in spiking neural networks},
  author  = {Kim, Robert and Sejnowski, Terrence J.},
  journal = {Nature Neuroscience},
  volume  = {24},
  number  = {1},
  pages   = {129--139},
  year    = {2021},
  doi     = {10.1038/s41593-020-00753-w}
}

@misc{bacvanski2025continuous,
title         = {Does Feedback Alignment Work at Biological Timescales?},
author        = {Bacvanski, Marc Gong and Ziyin, Liu and Poggio, Tomaso},
year          = {2026},
eprint        = {2510.18808},
archivePrefix = {arXiv},
primaryClass  = {cs.LG},
note          = {arXiv:2510.18808v2. Accessed 31/08/2026}
}

@inproceedings{bacvanski2026heteroassoc,
  author    = {Bacvanski, Marc Gong and Ziyin, Liu and Poggio, Tomaso},
  title     = {Continuous-Time Heteroassociative Memory at Biological Timescales},
  booktitle = {New Frontiers in Associative Memory Workshop at ICLR},
  year      = {2026},
  url       = {https://openreview.net/forum?id=WT2v0CXxp1}
}

@inproceedings{bacvanski2026ctkp,
  author    = {Bacvanski, Marc Gong and Ziyin, Liu and Poggio, Tomaso},
  title     = {Stabilizing continuous-time {K}olen--{P}ollack learning with a Scale-Balance Condition},
  booktitle = {High-dimensional Learning Dynamics Workshop at ICML},
  year      = {2026},
  url       = {https://openreview.net/forum?id=kmJuMuHe4j}
}

@article{francioni2026vectorized,
  title = {Vectorized instructive signals in cortical dendrites},
  author = {Francioni, Valerio and Tang, Vincent D. and Toloza, Enrique H. S. and Ding, Zilan and Brown, Norma J. and Harnett, Mark T.},
  journal = {Nature},
  year = {2026},
  doi = {10.1038/s41586-026-10190-7},
  volume = {652},
  pages  = {1254--1263}
}


\bibliographystyle{unsrtnat}
\arxivbibliography{pnas_ref_corrected}
\end{document}